\documentclass{article}

\usepackage{microtype}
\usepackage{graphicx}
\usepackage{subcaption}
\usepackage{booktabs}

\usepackage{setspace}
\usepackage{hyperref}

\usepackage{algorithm}
\usepackage[noend]{algpseudocode}
\algdef{SE}[SUBALG]{Indent}{EndIndent}{}{\algorithmicend\ }%
\algtext*{Indent}
\algtext*{EndIndent}
\algdef{SE}[SUBALG]{ForP}{EndForP}[1]%
{\textbf{for }#1 \textbf{do in parallel}}%
{}
\algtext*{EndForP}

\usepackage{xspace}
\newcommand{\photon}{Photon\xspace}

\usepackage[most]{tcolorbox}
\newtcolorbox{white}{colback=white!10!white,boxrule=0pt, top=0pt,bottom=0pt, left=0pt}
\newtcolorbox{blue}{colback=blue!10!white,boxrule=0pt, top=0pt,bottom=0pt, left=0pt}
\newtcolorbox{cyan}{colback=cyan!10!white,boxrule=0pt, top=0pt,bottom=0pt, left=0pt}
\newtcolorbox{red}{colback=red!10!white,boxrule=0pt,top=0pt,bottom=0pt, left=0pt}
\newtcolorbox{green}{colback=green!10!white,boxrule=0pt, top=0pt,bottom=0pt, left=0pt}

\colorlet{bluec}{cyan!20!white}
\colorlet{greenc}{green!20!white}

\makeatletter\def\Hy@Warning#1{}\makeatother

\usepackage{placeins}

\usepackage[capitalize]{cleveref}

\usepackage{paralist}

\usepackage{siunitx}

\usepackage{multicol, multirow}
\usepackage{xcolor,colortbl}
\definecolor{lightgray}{gray}{0.9} 
\definecolor{lightergray}{gray}{0.95}

\usepackage{eso-pic}
\usepackage{xparse}

\newlength{\badgewidth}
\newlength{\badgegap}
\newlength{\headerlogoheight}
\newlength{\headerlogogap}
\newcommand{\badgeList}{}
\newcommand{\headerLogoList}{}

\NewDocumentCommand{\addTopLeftLogo}{m}{%
\gappto{\headerLogoList}{\includegraphics[height=\headerlogoheight]{#1}\hspace{\headerlogogap}}%
}

\NewDocumentCommand{\addTopRightBadge}{O{} m}{%
\gappto{\badgeList}{\href{#1}{\includegraphics[width=\badgewidth]{#2}}\hspace{\badgegap}}%
}

\newcommand{\placeTopLeftLogos}{%
\AddToShipoutPictureBG*{%
\put(\LenToUnit{2.7cm},\LenToUnit{\paperheight - 2cm}){%
\makebox[0pt][l]{\headerLogoList}%
}%
}%
}

\newcommand{\placeTopRightBadges}{%
\AddToShipoutPictureBG*{%
\put(\LenToUnit{\paperwidth - 1.5cm - \badgewidth},\LenToUnit{\paperheight - 2cm}){%
\makebox[0pt][r]{\badgeList}%
}%
}%
}

\addTopLeftLogo{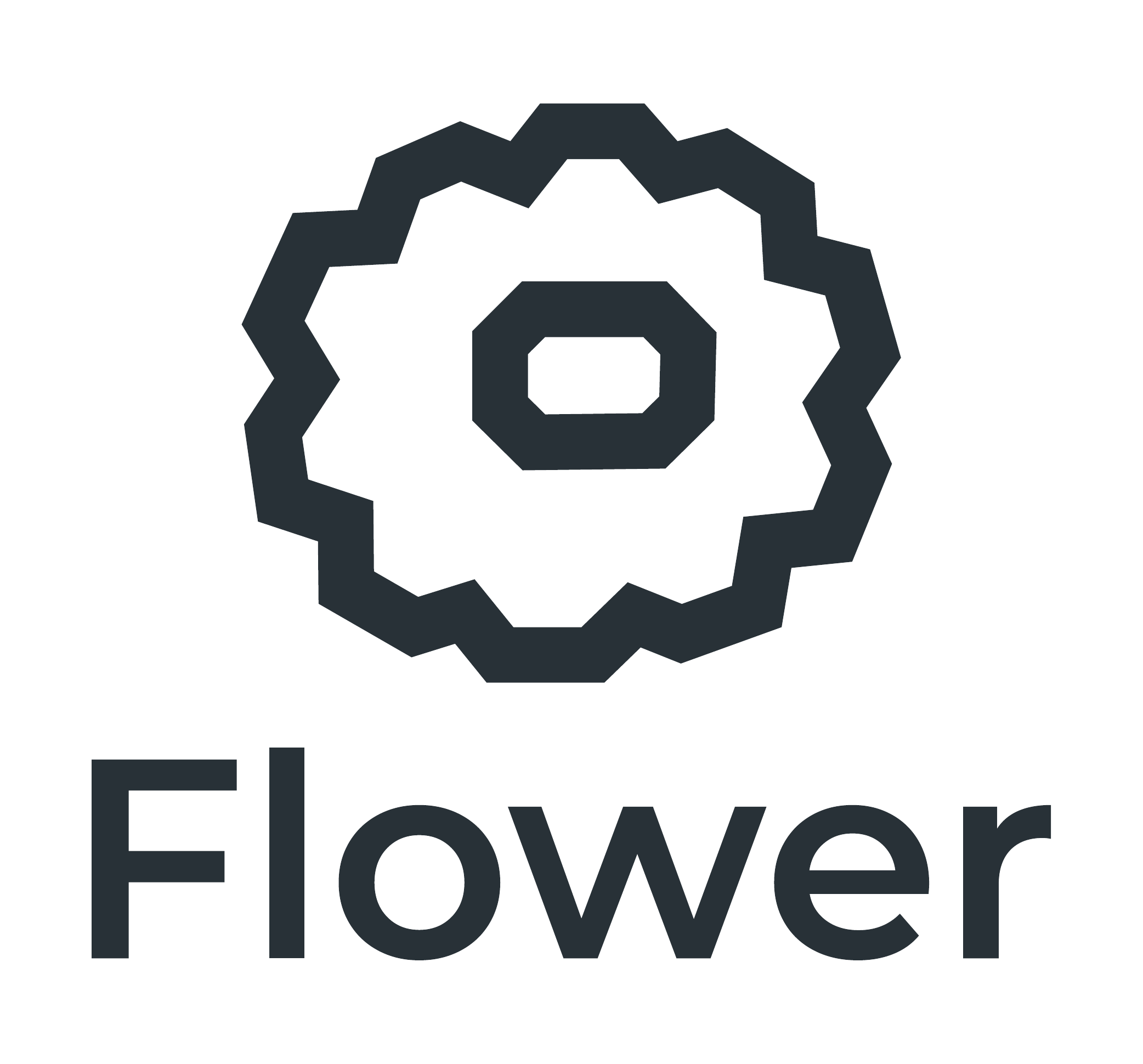}
\addTopLeftLogo{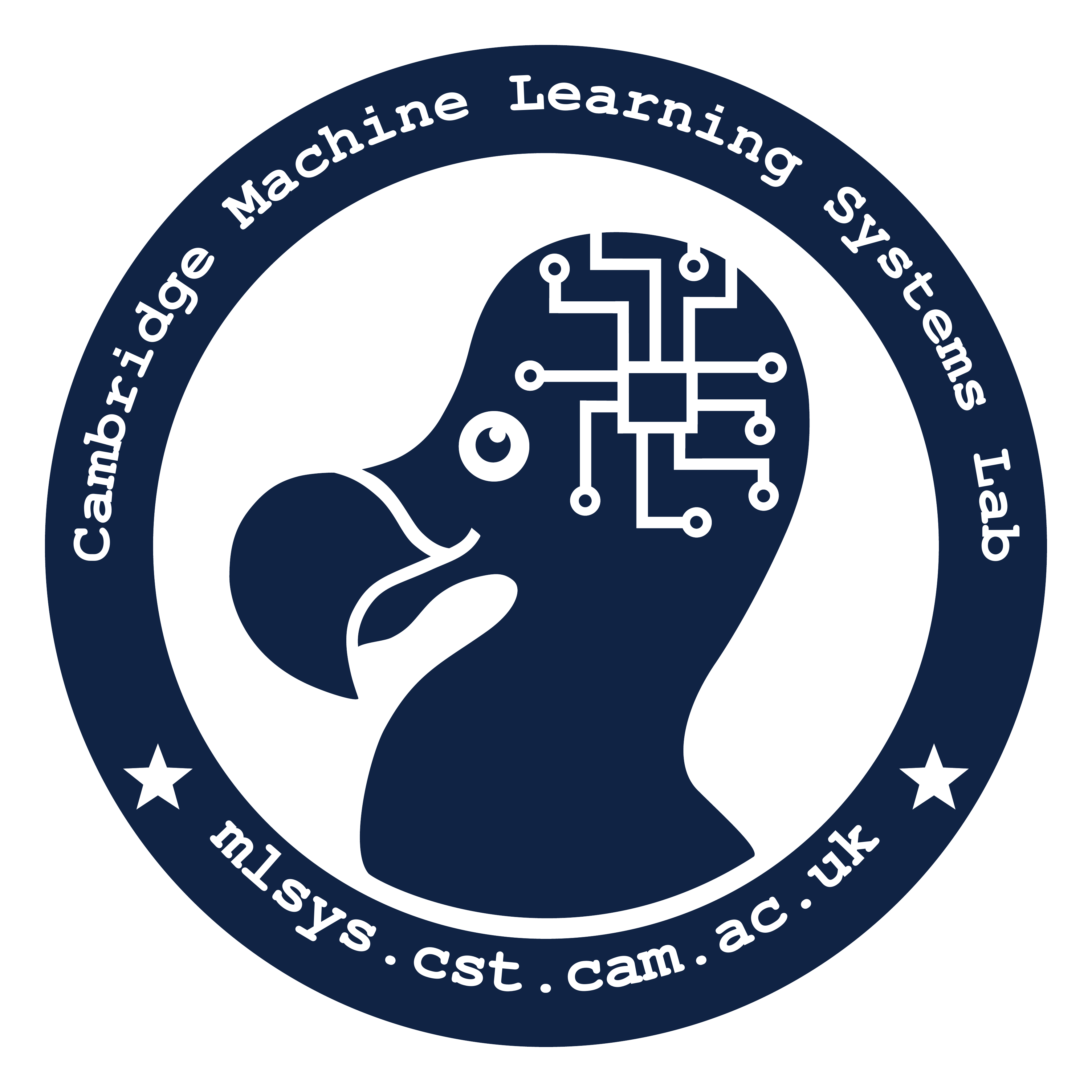}
\addTopLeftLogo{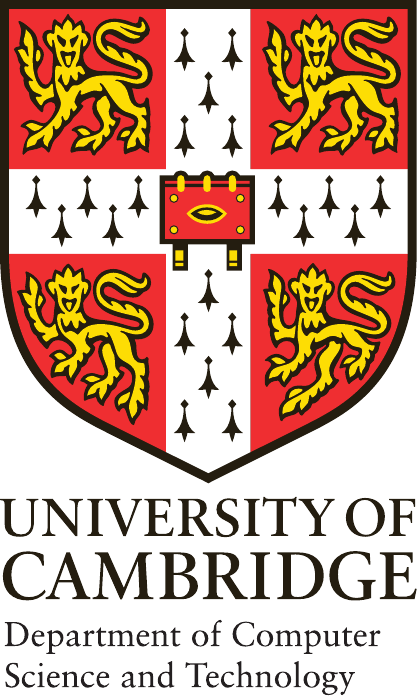}
\addTopRightBadge{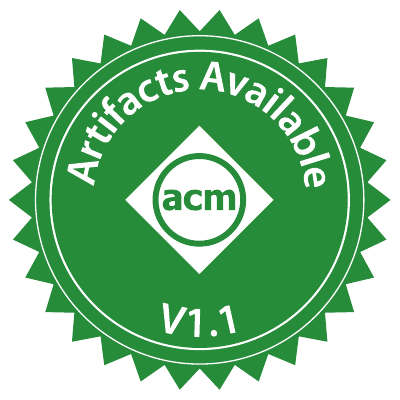}
\addTopRightBadge{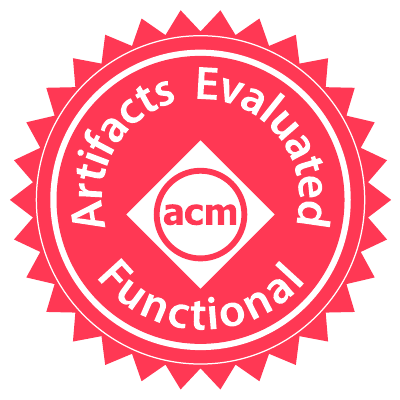}

\placeTopLeftLogos
\placeTopRightBadges

\usepackage{enumitem}
\usepackage[accepted]{mlsys2025}

\mlsystitlerunning{Photon: Federated LLM Pre-Training}

\begin{document}

\twocolumn[
\mlsystitle{Photon: Federated LLM Pre-Training}

\mlsyssetsymbol{equal}{*}

\begin{mlsysauthorlist}
\mlsysauthor{Lorenzo Sani}{cam,flwr}
\mlsysauthor{Alexandru-Andrei Iacob\textsuperscript{*}}{cam,flwr}
\mlsysauthor{Zeyu Cao\textsuperscript{*}}{cam}
\mlsysauthor{Royson Lee}{cam}
\mlsysauthor{Bill Marino}{cam}
\mlsysauthor{Yan Gao}{cam,flwr}
\mlsysauthor{Dongqi Cai}{cam,bupt}
\mlsysauthor{Zexi Li}{cam,zju}
\mlsysauthor{Wanru Zhao}{cam}
\mlsysauthor{Xinchi Qiu}{cam}
\mlsysauthor{Nicholas D. Lane}{cam,flwr}
\end{mlsysauthorlist}

\mlsysaffiliation{cam}{Department of Computer Science and Technology, University of Cambridge, Cambridge, United Kingdom}
\mlsysaffiliation{flwr}{Flower Labs}
\mlsysaffiliation{zju}{Zhejiang University}
\mlsysaffiliation{bupt}{Beijing University of Posts and Telecommunications}

\mlsyscorrespondingauthor{Lorenzo Sani}{ls985@cam.ac.uk}

\mlsyskeywords{Federated Learning, Distributed Learning, Systems for Machine Learning, Large Language Models}

\vskip 0.3in

\begin{abstract}
Scaling large language models (LLMs) demands extensive data and computing resources, which are traditionally constrained to data centers by the high-bandwidth requirements of distributed training. Low-bandwidth methods like federated learning (FL) could enable collaborative training of larger models across weakly connected GPUs or weakly connected clusters of GPUs if they can effectively be used for pre-training. Building robust low-bandwidth training systems can: (a) significantly reduce communication infrastructure costs, (b) minimize the impact of hardware failures, (c) widen the pool of usable GPUs, (d) enable collaborative training over the internet, and (e) allow dynamic compute sourcing based on factors like electricity prices.
Such advancements would lessen the dependence on specialized data centers, making large-scale AI training more accessible, cost-effective, and adaptable to real-time demands.
To achieve this, we introduce \photon, the first complete system for federated end-to-end LLM training, leveraging cross-silo FL for global-scale training with minimal communication overheads.
Using \photon, we train the first federated family of decoder-only LLMs from scratch.
We show that: (1) \photon can train model sizes up to $7$B in a federated fashion while reaching an even better perplexity than centralized pre-training; (2) \photon model training time decreases with available compute, achieving a similar compute-time trade-off to centralized; and (3) \photon outperforms the wall-time of baseline distributed training methods by $35\%$ via communicating $64\times$–$512\times$ less.
Our proposal is robust to data heterogeneity and converges twice as fast as previous methods like DiLoCo.
This surprising data efficiency stems from a unique approach combining small client batch sizes with extremely high learning rates, enabled by federated averaging's robustness to hyperparameters.
\photon thus represents the first economical system for global internet-wide LLM pre-training. \looseness=-1
\end{abstract}
]

\printAffiliationsAndNotice{\mlsysEqualContribution}

\section{Introduction}

Trends in developing state-of-the-art large language models (LLMs) suggest training ever-larger models on expanding datasets with growing compute resources \citep{OgScalingLaws, DBLP:conf/nsdi/JiangLZHCZPLXNJ24}.
The standard approach involves using a mix of distributed learning algorithms with high-bandwidth communication requirements deployed in single data center, e.g., distributed data parallelism across racks, pipeline parallelism across servers in a rack, and tensor parallelism across GPUs in a server \citep{10.1145/3394486.3406703, lee2024building, AlibabaHPN, llama3}.
Thus, increasing model size requires extending computing facilities to exploit high-bandwidth distributed training algorithms \citep{295545, DBLP:conf/icpp/YangPSWWWQP24}.

Recently, a small but growing interest in low-bandwidth distributed training algorithms has developed to exploit the worldwide distribution of computing facilities connected through the Internet \citep{DiLoCo, tang2024fusionllmdecentralizedllmtraining, DBLP:conf/nips/BorzunovRCBDBSR23, DBLP:journals/chinaf/MiXFWZZCL20, chang2023mining, sani2024futurelargelanguagemodel}.
If successful, low-bandwidth distributed training could overcome the need to build more extensive data centers.
The federated learning (FL) approach \citep{fedavg} is appealing as an additional layer of parallelism across poorly connected nodes, such as data centers distributed in different regions \citep{DiLoCo,distro,ScalingFedLLM,marfoq2020throughput} or internet-wide collaborative training.
The reasons for its appeal are three-fold: (1) FL optimizers derive from LocalSGD \citep{LocalSGD}, which allows more infrequent synchronization compared to distributed data parallelism, reducing demands on the communication infrastructure; (2) the size of text datasets makes it challenging to replicate across data centers \citep{mast}, which can be alleviated by bringing training to the data; and (3) federations scale seamlessly as participants join \citep{xu2024fwdllm}, i.e., as the total available compute expands, without needing to build additional costly infrastructure or reconfigure existing systems.

In this work, we present \photon, the first open-source FL system for executing pre-training of LLMs across a distributed setting - composed of individuals privately owing a handful of hardware accelerators - communicating through the Internet.
We show that \photon effectively navigates the trade-off between performance and efficiency and fills the gap for researchers and practitioners to federatedly pre-train high-performance LLMs off the shelf.
Notably, \photon has been used for pre-training the \emph{first} family of federated decoder-only large language models \emph{from scratch}, scaling model size up to 7B. Moreover, academic and industry researchers have used \photon to execute $1811$ experiments and submit six papers to international machine-learning venues.
We built \photon on the \texttt{Flower} framework -- the code is \href{https://github.com/relogu/photon}{publicly available}.

The contributions of this work are the following:
\begin{compactenum}
    \item We introduce \photon, the first open-source system for federated LLM pre-training over the Internet, enabling collaboration across private GPUs or distributed subsets of data centers worldwide. \photon has successfully trained the first federated family of decoder-only LLMs from scratch, reaching lower perplexities than centralized training for models up to $7$B parameters.
    \item We show that \photon achieves up to $20\%$ higher throughput (samples/sec) than centralized distributed training, requiring $64\times$–$512\times$ less communication. Furthermore, \photon is significantly more robust than standard data-parallel approaches since workers can continue training if another participant in the federation fails.
    \item We propose a novel federated pre-training approach exploiting the robustness to hyperparameters of federated averaging to combine small device batch sizes with high learning rates. This allows models trained with \photon to converge twice as fast as previous methods, such as DiLoCo~\citep{DiLoCo}.
    \item By combining high throughput with our optimization method, we demonstrate that the training time with \photon reduces as more compute resources are added up to a certain batch size limit. This achieves a compute-time trade-off similar to centralized pre-training.
\end{compactenum}

\photon enables scaling pre-training infrastructure to a global scale. Beyond allowing a potentially unprecedented number of GPUs to train a model effectively and robustly collaboratively, it could be used for novel applications. For example, distributing compute across nodes based on energy costs, carbon emissions, a fairness policy, or the proportion of unique private data.

\section{LLM pre-training under Decentralized System Conditions}
We focus on low-bandwidth distributed settings for training LLM.
In such scenarios, the standard distributed training method of synchronizing gradients at every batch step would incur very high overheads.
For example, the inter-worker communication costs of using distributed data parallelism (DDP) with \textit{Ring-AllReduce} \citep{Horovod} would be $\mathcal{O}(|\theta| \times T)$ where $|\theta|$ is the model size and $T$ is the number of training steps.
Moving to federated training allows us to reduce the communications costs \citep{LocalSGD,AdancesAndOpenProblems} to $\mathcal{O}(|\theta| \times \frac{T}{T_{\mathrm{local}}})$ by performing $T_{\mathrm{local}}$ steps on a ``worker'' prior to synchronization.

While highly beneficial from a communication perspective, infrequent synchronization significantly alters the optimization procedure \cite{LocalSGD, LocalSGD_Trade_Offs_At_Scale, DontUseLargeBatchesUseLocalSGD}.
Every gradient descent step in standard distributed training is executed on a fully up-to-date model.
At the same time, in federated learning, participants operate with stale parameters for $T_{\mathrm{local}}$ steps.
This poses challenges to reaching a similar level of data efficiency as standard distributed training \citep{field_guide_fedopt}.
In addition to this fundamental challenge, various FL settings can pose additional difficulties, such as hardware and data heterogeneity \citep{AdancesAndOpenProblems}. \looseness = -1

\subsection{Emerging Decentralized Scenarios}
Our design space has three particular settings for which low-bandwidth training methods are likely beneficial.

\textbf{Cross Data-center:} This scenario resembles standard distributed training, where multiple data centers collaborate to train models even larger than the current SOTA. Typically, distributed training approaches cannot operate over the low-bandwidth connection across data centers, forcing corporations to build ever-larger facilities.

\textbf{Cross-silo:} Here, we assume collaboration among several small organizations, each equipped with one to eight high-performance accelerators. In such cases, not only is the bandwidth across silos low, but silos may have an insufficient number of GPUs to saturate the batch-size requirements of even modest-sized models. 

\textbf{Collaboration via Commodity Hardware:} In this setup, individuals with a small number of consumer-grad GPUs collaborate in model training. This setting presents the above challenges in a harsher form due to strong VRAM constraints on commodity GPUs, making it difficult even to train a model without extreme CPU offloading.

Collaborating in such federated scenarios aims to benefit from \emph{more computing power and data sources} by \emph{achieving the machine learning (ML) optimization objective} quicker than what standard training can do in a single location.
Given our available resources, we focus on the standard cross-silo setting.
Thus, we \emph{assume} that every participant possesses the following minimal requirements: (a) one or many well-connected hardware accelerators, which can be sporadically available throughout a full training cycle; (b) sufficient memory to train the full model with a pre-defined small (local) batch size; (c) access to a pre-tokenized text corpus, either stored in the same facility or streamed through the Internet from a private data silo; and (d) a stable connection to the Internet with an average bandwidth of 2.5Gbps.

\subsection{Computation Efficiency}

LLM pre-training has presented many challenges to systems and architecture designers, as it has unprecedented memory footprints (VRAM) and requires extensive computing capabilities (FLOPs/s)~\citep{NEURIPS2022_c1e2faff}.
Most of these are mitigated by pooling extensive hardware accelerators and adopting distributed training algorithms.
Standard distributed training algorithms are based on $3$D parallelism, which applies data parallelism (DP) across racks in a data center~\citep{NIPS2012_6aca9700}, pipeline parallelism (PP) across servers in a rack~\citep{PipeDream}, and tensor parallelism (TP) across GPUs in a server~\citep{megatron-LM}.
The common practice is designing computing facilities to fully exploit $3$D parallelism for optimal resource utilization.

Achieving optimal resource utilization requires thoughtful configuration of the hyperparameters and the mixture of distributed algorithms, which becomes more challenging as the scale increases.
This tuning involves choosing the most appropriate batch size that will result in the least expensive gradient accumulation (ideally, none).
We assume the participants in the distributed settings discussed here thoroughly understand their available hardware and their interplay with the $3$D parallelism of the target model size.
We construct our evaluation on settings running full batch steps matching their resources without any gradient accumulation.

\subsection{Sourcing and Moving Data}

ML workloads, such as LLM pre-training, require massive training data and computational resources, naturally distributed across several regions.
To achieve reasonable efficiency, ML infrastructures are designed to follow the \textit{data-GPU collocation} principle, i.e., data warehouses are colocated in the same region with GPU clusters to avoid relying on the cross-region network bandwidth (usually $10$ times lower than intra-region network bandwidth).
Large-scale infrastructures face the challenge of satisfying \textit{data-GPU collocation} for \textit{exabytes} of data, tens of regions with thousands of GPUs.
Data is also continuously produced and removed, making the collocation task even more challenging.
A worldwide scheduler, such as MAST \citep{mast}, optimizes the cross-regional data placement daily, leveraging algorithms that can take up to $5$ hours to complete their task.
In this context, our work tackles a setting, usually referred to as \textit{training-at-home}.
Our \photon takes advantage of the available computing power at clients, where data is stored, resulting in the following benefits: (a) it doesn't require particular data placement optimization; (b) it can leverage low-hanging fruit local storage optimizations, such as data pre-tokenization, (c) it is compliant with privacy constraints as it doesn't move data.

\subsection{Cross-silo Communication}

Training procedures based on $3$D parallelism leverage high-bandwidth networks supported by \textit{intra-datacenter} networking solutions such as \texttt{RoCE} and \texttt{InfiniBand} with the typical link speed from $100$Gbps up to $400$Gbps per link. \citep{DBLP:journals/corr/abs-1802-05799, 10.1145/3651890.3672233, 10.1145/3663408.3663409}.
Their communication efficiency of the standard distributed training approach is heavily impacted by slow network links, which makes them unsuitable for \textit{cross-region} applications we are interested in as their network bandwidth ranges from $0.8$Gbps to $40$Gbps.
In this work, we use distributed training algorithms based on LocalSGD, which requires less frequent communication across workers \citep{LinearSpeedupCommsEfficientSGD} and strongly reduces the overheads of possessing slower links across workers.
\section{Architecture and Design of \photon}
\begin{figure}[t]
    \centering
    \includegraphics[width=\columnwidth]{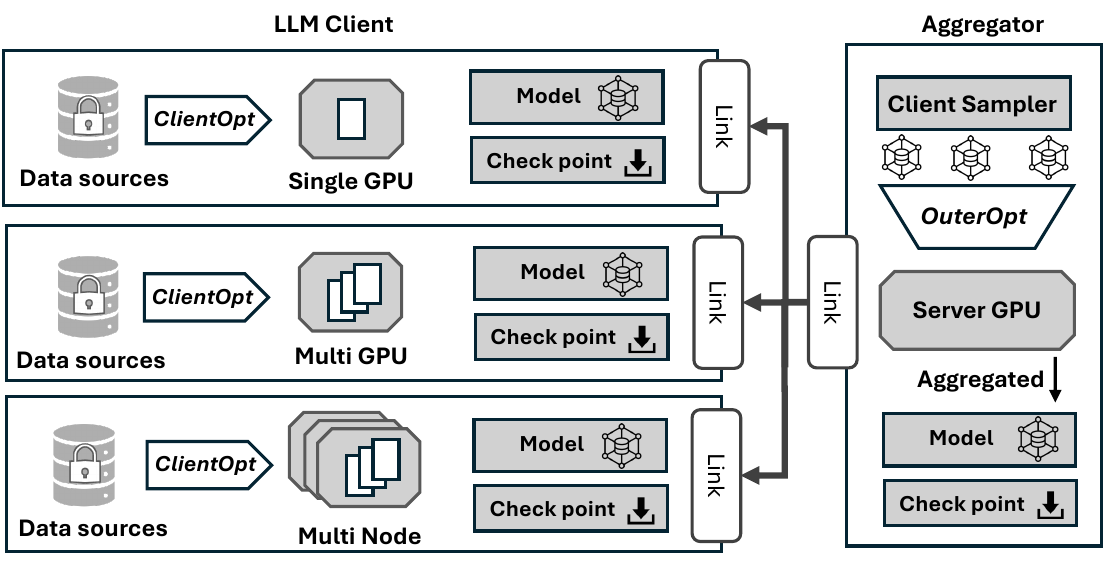}
    \caption{\textbf{Systematic diagram of \photon's three principal components - \photon aggregator, \photon LLM clients, and \photon data sources.} Arrows describe interactions and message exchanges. The \photon aggregator can only communicate with the \photon LLM nodes through the \photon link. The instances responsible for storing the data samples, the \photon data sources, can uniquely stream to the \photon client bound to them.}
    \label{fig:diagram:systemx-1}
\end{figure}

To enable collaborative and effective cross-silo FL pre-training of LLMs with limited inter-data center communication, \photon follows three core principles: \textit{broad inclusivity of data and compute sources}, \textit{minimal compute requirements}, and \textit{scalable local training pipelines}.
These principles maximize client resource utilization and ensure a robust design.
We now present \photon’s architecture, beginning with a brief overview of its core innovations.
The main components are summarized in \Cref{fig:diagram:systemx-1}.

\textbf{Adaptive Local Parallelism:} \photon integrates standard distributed training techniques with federated learning, optimizing training data storage, transfer, parameter communication, and aggregation.
It adapts to each client’s connectivity and topology, allowing automatic selection between standard distributed training and low-bandwidth LocalSGD.

\textbf{Improved Model Generalization:} The federated optimization we adopt produces robust model minima and is resilient to hyperparameter variations due to noise injection~\citep{DontUseLargeBatchesUseLocalSGD} and meta-learning effects~\citep{REPTILE}, ensuring convergence across varied client participation levels, data heterogeneity, and local training hyperparameters. Since robust model minima, defined as model parameters whose loss does not significantly change upon perturbation, are known to generalize better and produce lower validation loss~\citep{SharpMinimaLargeBatch,DontUseLargeBatchesUseLocalSGD}, this partially explains the outperformance that models trained with \photon show over centralized pre-training.

\textbf{Exploiting Small Batches and High Learning Rates:} \photon's robustness to hyperparameter choices enables the use of \textbf{small} (hardware-determined) local batch sizes, which promote flat minimizers of the loss that generalize better~\citep{SharpMinimaLargeBatch}, along with \textbf{high learning rates} decayed over an extended period—typically unstable with small batches—allowing us to maintain data efficiency.
For example, if centralized training uses a decay period $T$ with batch size $\mathcal{B}$, federated learning enables us to extend it to \( T \times \frac{\mathcal{B}}{\mathcal{B}_{\mathrm{small}}} \). In our experiments, using small batch sizes $\mathcal{B}_{\mathrm{small}}$ in centralized training always resulted in model divergence unless the maximal learning rate was reduced linearly w.r.t the batch size. Full details are in \cref{app_subsec:fl_llm_pretrain}.

\subsection{Architecture}

\photon consists of the following core components.

\noindent
\textbf{Aggregator (Agg):} Agg serves as the central server orchestrating the federated training process.
At the start of each round, it activates the \textit{Client Sampler} to access and select LLM-C instances according to the optimization algorithm’s requirements.
Agg then uses Link to relay messages between LLM-C clients.
Once results are received from LLM-C, \texttt{OuterOpt} aggregates updates and applies the optimization to the global model, followed by checkpointing.

\noindent
\textbf{LLM Client (LLM-C):} LLM-C is the distributed client in \photon responsible for the local training pipeline within the federated optimization process.
Each LLM-C can connect to Agg at any point during training.
The \texttt{ClientOpt} trains the model received from Agg on local data, utilizing various distributed algorithms suited to each LLM-C's hardware capabilities.
Model updates and training metadata are then exchanged with Agg through Link.
The training state is regularly checkpointed for fast recovery in case of failure.

\noindent
\textbf{Data Sources (DS):} Data Sources serve as the data storage for \photon, meeting the federated learning requirements regarding data location and exchange protocols.
Each private DS is linked to an individual LLM-C within the same client’s data domain, generating a continuous data stream that matches \texttt{ClientOpt}'s training throughput.
This decoupled structure allows institutions with large data silos to obtain paired computation through LLM-C without sharing data globally.
Additionally, public DS can be configured for data sharing among LLM-C clients to support collaboration or data-sharing agreements between participants.

\subsection{Operation}

\begin{figure}[t]
    \centering
    \vspace{0.3cm}
    \includegraphics[width=0.9\columnwidth]{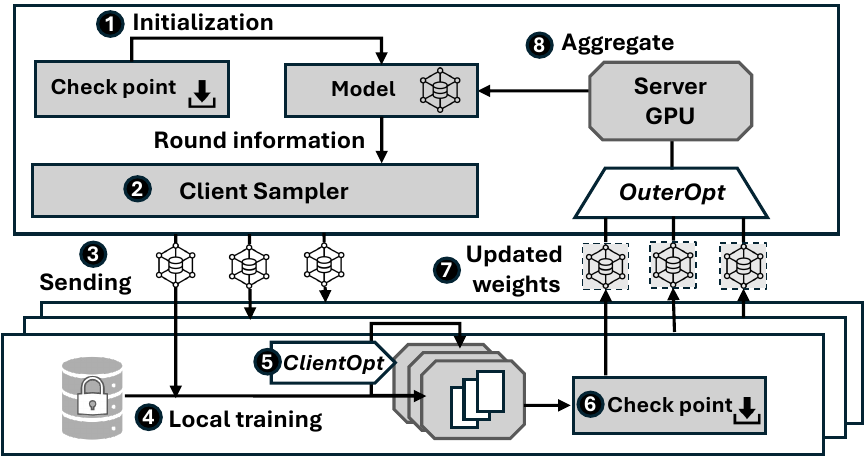}
    \caption{\textbf{Information flows between LLM clients, the aggregator, and data sources.} Following initialization (1), An LLM client selected by the client sampler (2) receives model parameters from the server (3), trains on data from a data source (4,5), checkpoints (6), and then returns the updated parameters (7) to the aggregator for federated optimization (8).}
    \label{fig:diagram:systemx-2}
\end{figure}

We detail \photon's workflow, as illustrated in Algorithm~\ref{alg:systemx-pipeline}, and present a workflow visualization in \Cref{fig:diagram:systemx-2}.

\photon assumes collaboration among multiple independent institutions, each with distinct data and compute silos, to pre-train LLMs with their own LLM-C.
For enhanced data protection, the Aggregator (Agg) can be hosted by one of the institutions or a trusted third party.
At the start of training, Agg initializes the model or sources it from one of the LLM-C~(\colorbox{bluec}{\small L.$2$}).
After initialization, Agg coordinates round information for client sampling~(\colorbox{bluec}{\small L.$3-4$}), sends the model to each sampled client, and collects results~(\colorbox{bluec}{\small L.$5-7$}).
It then aggregates the local models to construct a new federated model~(\colorbox{bluec}{\small L.$8-10$}), checkpointing the global model at the end of each round to enhance robustness~(\colorbox{bluec}{\small L.$11$}).

At each round, sampled LLM-C instances initiate their local training pipeline by acquiring data streams from their respective private DS~(\colorbox{greenc}{\small L.$14$}).
For enhanced evaluation, \photon allows DS to use public data sources when configured.
Each LLM-C evaluates available hardware resources to select the optimal execution strategy~(\colorbox{greenc}{\small L.$15$}), either utilizing fast interconnection between GPUs~(\colorbox{greenc}{\small L.$16-18$}) or applying an additional level of federated optimization if inter-GPU connectivity is limited~(\colorbox{greenc}{\small L.$19-23$}).
In the latter case, LLM-C performs an extra level of local aggregation~(\colorbox{greenc}{\small L.$24-25$}).
A local checkpoint is also maintained for quick recovery in case of failure~(\colorbox{greenc}{\small L.$27$}).
After completing the local training pipeline, LLM-C applies post-processing (e.g., gradient clipping, compression, or differential privacy noise injection) before returning updates to Agg~(\colorbox{greenc}{\small L.$28$}).

\begin{algorithm}[H]
\caption{\photon execution pipeline} \label{alg:systemx-pipeline}
\small
\begin{onehalfspace}
\begin{algorithmic}[1]
\Require{Number of rounds $T$, training population $P$}
\Require{Number of clients per round $K$, hyperparameters $H$}
\begin{cyan}
\Procedure{Aggregator}{$T$, $K$, $H$, $P$}
    \State{$\theta^0 \gets \mathtt{InitModel}(H)$}
    \For{each round $t=1,2,3,\ldots,T$} 
    \Indent
        \State{$\mathcal{C} \sim \mathcal{U}(P, K)$} 
        \ForP{$k \in \mathcal{C}$} 
            \State{$\theta_k^t, \mathcal{M}^t_k \gets \Call{LLM Client}{k, \theta^t, H}$}
            \State{$\Delta^{t}_{k} \gets \theta^t - \theta_k^t $}
        \EndForP
        \State{$\Delta^t \gets \frac{1}{|C|}\sum_{k\in C}{\Delta^{t}_{k}}$} 
        \State{$\theta^{t+1} \gets \mathrm{ServerOpt}(\theta^t, -\Delta^t, t)$}   
        \State{$\mathcal{M}^{t+1}_k \gets \mathtt{AggMetrics}(\mathcal{M}_k^t | \forall k \in \mathcal{C})$} 
        \State{$\mathtt{Checkpoint}(\theta_k^{t+1})$} 
    \EndIndent
    \State{\Return{$\theta_k^{T+1}$}}
    \EndFor
\EndProcedure
\end{cyan}
\begin{green}
\Procedure{LLM Client}{$k, \theta^t, H$}
        \State{$\mathcal{D}_k \gets \mathtt{BindStream}(k)$}
            \State{$I_k \gets \mathtt{GetNodes}(k)$} 
            \If{$\mathtt{HasRDMA}(I_k)$}
                \State{$B_k \gets \mathtt{CalcBatchSize}(I_k)$ } 
                \State{$\theta^t_k, \mathcal{M}_k^t \gets \mathtt{TrainClient}(\theta^t, \mathcal{D}_k, B_k, H)$} 
            \Else
                \ForP{node $i \in I$}  
                    \State{$B_k^i \gets \mathtt{CalcBatchSize}(i, I_k)$ }
                    \State{$\mathcal{D}_k^i \gets \mathtt{PartitionStream}(i, \mathcal{D}_k)$} 
                    \State{$\theta^t_i, \mathcal{M}_i^t \gets \mathtt{TrainClient}(\theta^t, \mathcal{D}_k^i, B_k^i, H)$} 
                \EndForP
                 \State{$\theta^t_k \gets \frac{1}{|I|}\sum_{i\in I}{\theta^{t}_{i}}$} 
                 \State{$\mathcal{M}^t_k \gets \mathtt{AggMetrics}(\mathcal{M}_i^t | \forall i \in I_k)$} 
            \EndIf
            \State{$\mathtt{Checkpoint}(\theta_k^t, \mathcal{D}_k)$} 
            \State{$\theta_k^t \gets \mathtt{PostProcess}(\theta_k^t, \mathcal{M}_k^t)$} 
            \State{\Return{$\theta_k^t, \mathcal{M}^t_k$}}
\EndProcedure
\end{green}
\end{algorithmic}
\end{onehalfspace}
\end{algorithm}

\section{\photon Implementation}\label{sec:systemx_implementation}

The \photon implementation is highly optimized for the unique challenges of federated pre-training.
Thus, it has the following objectives: (a) allow for efficient, intermittent data transfer between clients, data sources, and the aggregator; (b) effectively select the distributed training algorithm for a given client given their local GPU topology; and (c) exploit the federated communication topology to select the fastest aggregation method.
We now discuss how our implementation, consisting of approximately \num{16273} lines of code, achieves these aims.

\noindent
\textbf{Link between Agg and LLM-C:}
To enable efficient communication between Agg and LLM-C, \photon includes a dedicated communication module, \textit{Link}.
This module assumes a relatively fast and stable internet connection of at least 1Gbps between Agg and LLM-C, which is appropriate for the cross-silo federated setting.
Serving as the communication gateway, Link uses secure TLS encryption and supports secure aggregation~\citep{SecAggOG} for enhanced privacy, if needed.
Beyond model updates, message payloads carry metadata, including training and evaluation instructions, metrics, and global instructions. Link provides an extensible post-processing pipeline by leveraging model compression and pruning techniques.
By default, \photon uses lossless compression techniques without pruning.

\noindent
\textbf{Data Streaming for DS:}
Data is naturally distributed in cross-silo settings; therefore, our DS implementation breaks from the traditional one-to-one mapping between compute and data resources.
Treating data as streams from one or multiple DS elements, \photon enables mixing arbitrary data streams with precise control over sampling across such streams.
This decoupling allows data providers to operate independently of compute providers, broadening our federation.
Moreover, this design reduces core network utilization if a DS can communicate with an LLM-C over a separate network.
To optimize data streaming, \photon DS employs caching alongside optional data pre-tokenization and compression.
These optimizations heuristically minimize storage overhead during the transfer from data producers to data consumers, reduce compute demands on data consumers, and maintain the throughput required by each LLM-C.

\noindent
\textbf{Optimal Training Strategy Selection for LLM-C:}
To accommodate the heterogeneous nature of the hardware, \photon’s LLM-C can support a wide range of configurations, provided the model fits within the available total VRAM with a batch size of at least one sample.
An LLM-C hardware setup can include a single GPU, multiple GPUs within a server node, or multiple servers connected by high-bandwidth interconnects.
\photon aims to maximize throughput for each LLM-C by selecting an optimal training strategy through a heuristic-based approach that is summarized as follows:
\begin{compactenum}
 \item If a model and sufficient batch size fit within a single GPU, enabling the client to keep pace with the federation, LLM-C assigns a dedicated GPU to each client. \looseness=-1
 \item For nodes with multiple GPUs, LLM-C uses either DDP or FSDP, depending on whether a model with a viable batch size fits within a single GPU.
 \item When clients have a cluster of GPU-equipped machines, LLM-C selects a strategy based on the cluster interconnection speed. High-bandwidth interconnects lead to DDP or FSDP, similar to the previous variant. In contrast, lower bandwidth may necessitate constructing sub-federations with further data sub-partitioning, with each partition trained independently.
\end{compactenum}
In each case, the strategy ensures optimal compute utilization within the LLM-C topology.
All LLM-C training strategies are transparent to Agg, enhancing the federation's scalability and extensibility.
Our experiments primarily use common cross-silo settings, such as allocating one GPU per client, reflecting typical institutional bandwidth constraints between servers.
However, as demonstrated later, \photon is flexible and supports better-interconnected topologies.

\noindent
\textbf{Topology Between Clients:}
For \photon to efficiently pre-train an LLM in a federated setting, parameter aggregation must be effectively managed within current communication constraints.
In general, a model with size $M = |\theta|$ trained across $N$ workers can be aggregated in three variants, each with other distinct use cases, advantages, and limitations: 
\begin{compactenum}
    \item A \textit{parameter server} (PS) receives all updates from participating workers. This variant is ideal for relatively small $N$, as the data received by the server scales in $\mathcal{O}(N M)$. It handles worker dropouts well by providing a partial update derived from surviving workers and is the only viable option when privacy restrictions prohibit peer-to-peer communication. 
    \item Workers communicate directly via \textit{AllReduce} (AR). In this setup, each worker sends its model to all peers and receives models from all others, resulting in $\mathcal{O}(N^2 M)$ data transmission per worker. Like PS, AllReduce tolerates dropouts well, enabling partial updates from remaining workers; however, privacy limitations may restrict peer-to-peer communication.
    \item Workers use \textit{Ring-AllReduce} (RAR), communicating over a ring topology for efficient aggregation. This bandwidth-optimal method requires each worker to send/receive $\mathcal{O}(M)$ data, with the bottleneck being the slowest link in the ring. RAR does not tolerate dropouts and has similar privacy considerations to AllReduce.
\end{compactenum}
Each method has its constraints, and \photon adapts to select the most efficient option for each scenario.

\section{Evaluation}

\begin{figure}[t]
    \centering
    \includegraphics[width=\columnwidth]{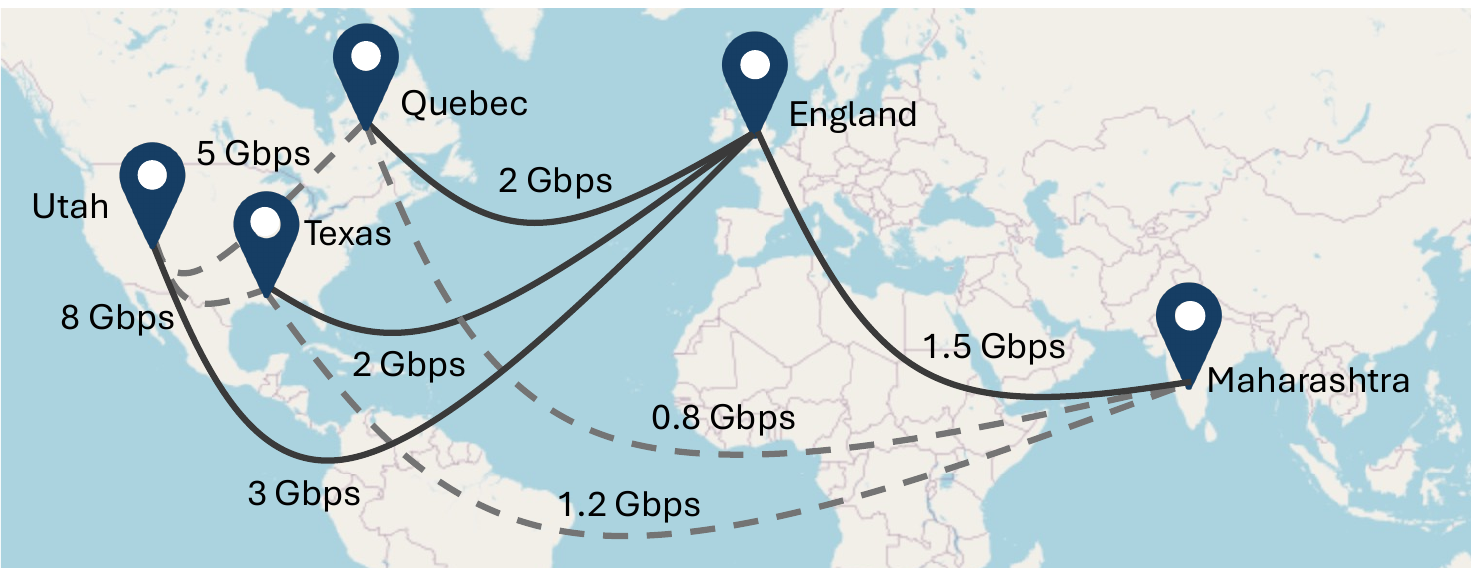}
    \caption{\textbf{The locations and bandwidth of participating clients in the Federation}, with multiple nodes equipped with H100s at each site. More details are available in \cref{tab:regions}. Bandwidth between regions varies significantly, impacting the efficiency of \photon's aggregation procedures. The map shows the RAR topology (gray dashed line) and the PS topology (black solide line). The slowest link in the RAR topology, between Maharashtra and Quebec, acts as a bottleneck. In the PS topology, the connection speed to England limits each update's communication. }
    \label{fig:world-map-bandwidth}
    \vspace{-0.2cm}
\end{figure} 

\begin{table}[t]
\caption{\textbf{Computational resources of different regions.} 
For each region, ``num. of clients \textbf{x} num. of GPUs held by each client'' is shown, e.g., 1 x 8 H100. \photon enabled this globally distributed setup, overcoming challenges of low average bandwidth.}\label{tab:regions}
\centering
\resizebox{\columnwidth}{!}{
\begin{tabular}{@{}ccccccc@{}}
\toprule
\textbf{Size} & \textbf{Agg} & \textbf{England} & \textbf{Utah} & \textbf{Texas} & \textbf{Quebec} & \textbf{Maharashtra} \\ \midrule
\rowcolor{lightgray} 7B & England & - & 1 x 8 H100 & 1 x 8 H100 & 1 x 8 H100 & 1 x 8 H100 \\
\rowcolor{lightergray} 3B & England & - & 1 x 4 H100 & 1 x 4 H100 & 1 x 4 H100 & 1 x 4 H100 \\
\rowcolor{lightgray} 1B & England & 1 x 2 H100 & 2 x 2 H100 & 2 x 2 H100 & 2 x 4 H100 & 
1 x 4 H100 \\
\rowcolor{lightergray} 125M & England & 2 x 1 H100 & 2 x 1 H100 & 2 x 1 H100 & 2 x 1 H100 & 2 x 1 H100 \\ \bottomrule
\end{tabular}
}
\vspace{-0.3cm}
\end{table}

This section demonstrates the effectiveness of \photon in training large language models (LLMs) from scratch in federated settings, compared to models trained using standard centralized methods and prior low-bandwidth distributed approaches.
We evaluate models based on final performance, compute efficiency, and communication scalability.

\subsection{Experimental Setup}\label{sec:exp_setup}

Our experiments employ a cross-silo FL setup where clients are equipped with one or more high-end GPUs, such as Nvidia H100s, and are interconnected via the Internet.
The computational resources used on each client vary based on model size: million-parameter models are trained using a single GPU, whereas billion-scale models may require up to $8$ GPUs per client. We present the five locations used during training for our distributed setup in \cref{fig:world-map-bandwidth}, with details on the number of clients and GPUs available per client provided in \cref{tab:regions}.\looseness=-1

The client's local batch size is determined by its VRAM, model size, and optimal throughput, leveraging heuristics similar to those proposed by the Microsoft DeepSpeed AutoTuner \citep{microsoft_deepspeed, deepspeed_autotuner}.
For instance, clients training a $125$M parameter model use $1$ Nvidia H100, processing a hardware-determined local batch size $B_l = 32$, without gradient accumulation or activation checkpointing, to maximize throughput. 
Given the assumption of homogeneous client resources, all clients independently employ the same local throughput optimization strategies.
Experiments are conducted using PyTorch (v2.4.0) with mixed precision context \texttt{bfloat16} (BF16).
The model architecture is based on the MPT (Mosaic Pre-Trained) family of decoder-only transformers \citep{mpt_blogpost}, with the MPT centralized training recipe used as a baseline configuration.\looseness=-1

\textbf{Datasets and Training Recipe.} We model data distribution across clients by randomly partitioning the \texttt{C4} \citep{C4} dataset uniformly into $64$ equally sized shards.
$N$ clients, hence, refer to a subset of $N$ shards from these $64$ shards.
The local training step for clients uses AdamW \citep{AdamW} as the optimizer, with a cosine learning rate schedule and an initial linear warm-up phase. We report all hyperparameters in \cref{app_sec:hypeparams}.
The learning rate schedule varies by model size and is adjusted to match the batch size and total token count seen in centralized training.
We experiment with different numbers of local steps per round, specifically $\{62, 128, 512\}$, which defines the amount of local work done by clients. 
Model performance is evaluated using perplexity on the full \texttt{C4} validation set.

To explore the robustness of \photon to data heterogeneity, we also explore a setting where clients hold data from a variety of text sources representing diverse categories.
Specifically, we use The Pile~\citep{ThePile} dataset, which includes text sources from \texttt{ArXiv} (academic), \texttt{C4} and \texttt{Wikipedia} (internet), and \texttt{Project Gutenberg} (prose) to capture a range of language styles while providing sufficient data for scaling.
For our experiments with full participation, we explore three configurations: four clients (one text source per client), eight clients (two text sources split into two clients), and sixteen clients (four text sources split into four clients).
For partial participation, we adopt the sixteen-client configuration, sampling $25\%$, $50\%$, and $100\%$ of clients per round, with evaluation on \texttt{C4}.

\begin{figure}[t]
    \centering
    \includegraphics[width=0.45\columnwidth]{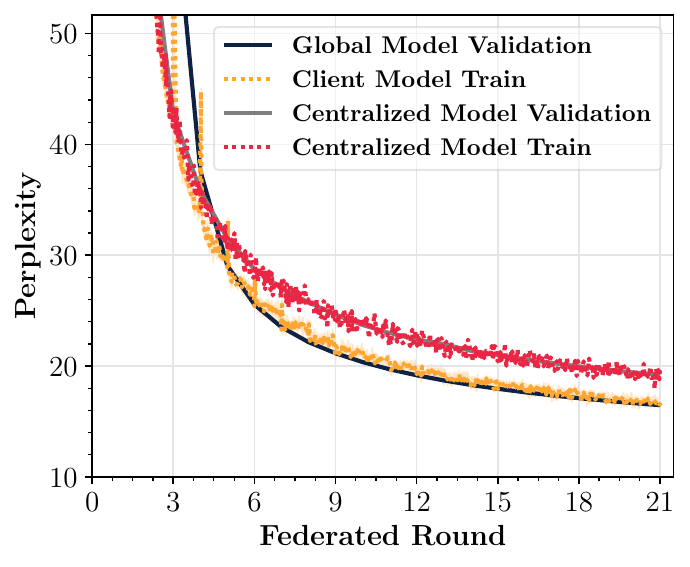}
    \includegraphics[width=0.45\columnwidth]{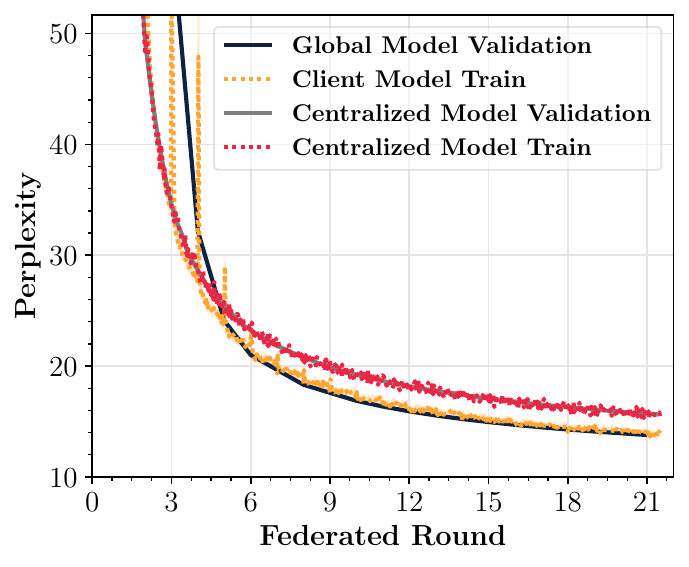}
    \caption{\textbf{Comparison of perplexity convergence (\textit{lower is better}) for \photon and centralized training with $3$B (left) and $7$B (right) models.} The federated global model was evaluated on the \texttt{C4} test set, with averaged train perplexities across clients and centralized train/test perplexities presented for both models. These large federated models show lower perplexity than centralized models and remain stable during aggregation, with minimal perplexity spikes after early rounds.}
    \label{fig:fed:perplexity_3b_and_7b}
    \vspace{-0.3cm}
\end{figure} 

\subsection{\photon trains Billion-sized LLMs}
 
Using \photon, we train models up to $7$B parameters from scratch in federations of $16$ to $4$ clients, using full participation every round and prove they outperform their centralized counterparts. 
While previous results suggest that federated optimization is generally inferior to standard SGD in terms of loss~\citep{field_guide_fedopt} and that LLM pre-training may be less sample-efficient~\citep{DiLoCo,sani2024futurelargelanguagemodel}, we find it to be quite competitive in terms of wall-clock time and, comparable if not slightly superior in terms of loss.
As shown in \cref{fig:fed:perplexity_3b_and_7b,tab:perplexity_imp}, the $3$B and $7$B models achieve $13.8\%$ to $16.9\%$ lower perplexity than models trained using centralized training.
As shown in \cref{tab:federation_scalability}, such models can perform similarly in terms of perplexity after sufficient training.
For instance, when utilizing identical computational resources with a 10 Gbps connection and Ring-AllReduce, \cref{tab:federation_scalability} demonstrates that federated optimization for the 7B model to reach the same perplexity requires only $95.6$ total hours whereas centralized methods require over 147 hours.
Despite the compute time of the $7$B model being almost $2\times$ higher, the $500\times$ communication reduction of \photon can bring an overall reduction in wall time.
Higher-bandwidth connections such as \texttt{InfiniBand} would bring the total time closer to the compute time. Moving beyond perplexity, we also show that the models produced by \photon are effective for downstream tasks in \cref{app_sec:downstrea_eval}.

\begin{table}[t]
\small
\caption{\textbf{Our results show federated models to be comparable to centralized and potentially superior as they obtain lower perplexity (PP) given the same computational resources.} Their perplexity gains grow with model size.}\label{tab:perplexity_imp}
\centering
\begin{tabular}{@{}lccc@{}}
\toprule
\textbf{Size} & \textbf{Fed PP} & \textbf{Cent PP} & \textbf{Gain ~(\%)} \\ \midrule
\rowcolor{lightgray} \rowcolor{lightergray} $1.3\text{B}$ & $\mathbf{20.1}$ & $23.2$ &  $\mathbf{13.4}\%$ \\
\rowcolor{lightgray} $3\text{B}$ & $\mathbf{15.7}$ & $18.2$ & $\mathbf{13.7}\%$ \\
\rowcolor{lightergray} $7\text{B}$ & $\mathbf{13.8}$ & $16.6$ & $\mathbf{16.9}\%$ \\ \bottomrule
\end{tabular}
\vspace{-0.3cm}
\end{table}

We observe that the perplexity gap between federated and centralized models widens at larger scales; smaller models converge to similar performance levels across methods.
Additionally, as model size grows, training stability improves, as we observe perplexity spikes decreasing in magnitude compared to smaller models.
Our observed performance improvements stem from the noise-injecting~\citep{DontUseLargeBatchesUseLocalSGD} and meta-learning~\citep{REPTILE} of our federated optimization.
As we will later see, federated optimization supports high learning rates with small batch sizes, enhancing generalization via noise injection~\citep{SharpMinimaLargeBatch}.

\begin{table}[t]
\caption{\textbf{We report system metrics for the billion-scale models trained with \photon and their centralized baselines}, including total wall time with compute and communication time breakdowns. We provide GPU efficiency metrics such as average GPU utilization and corrected Model FLOPs Utilization (MFU) during local computation. While federated optimization may increase compute time, \photon’s federated approach shortens overall training time by reducing communication steps, assuming consistent setups: the number of federated clients matches data-parallel workers, aggregation uses \textit{Ring-AllReduce} with a fixed 10Gbps bandwidth for the slowest link, and both clients and workers maintain the same mini-batch throughput.}\label{tab:federation_scalability}
\vspace{0.1cm}
\resizebox{\columnwidth}{!}{
\begin{tabular}{@{}lccccc@{}}
\toprule
\multirow{2}{*}{\textbf{Model}} & \textbf{Wall} & \textbf{Compute} & \textbf{Comm.} & \textbf{Local Util.} & 
\textbf{Local MFU} \\
& \textbf{Time [h]} & \textbf{Time [h]} & \textbf{Time [h]} & \textbf{GPU [$\%$]} & \textbf{per device}\\
\midrule
\rowcolor{lightergray}
Cen-$1.3$B  & $26.7$ $(1\times)$      & $6.5$ $\boldsymbol{(1\times)}$       & $20.2$ $(1\times)$      & $74$ & $0.2531$ \\
\rowcolor{lightgray}
Fed-$1.3$B  & $18.02$ $\boldsymbol{(0.67\times)}$  & $18.0$ $(2.8\times)$    & $0.02$ $\boldsymbol{(0.001\times)}$  & $83$ & $0.3546$ \\
\midrule
\rowcolor{lightergray}
Cen-$3$B    & $56.6$ $(1\times)$      & $16.1$ $\boldsymbol{(1\times)}$      & $40.48$ $(1\times)$     & $81$ & $0.051$  \\
\rowcolor{lightgray}
Fed-$3$B    & $25.2$ $\boldsymbol{(0.45\times)}$   & $25.1$ $(1.6\times)$    & $0.05$ $\boldsymbol{(0.001\times)}$   & $78$ & $0.076$  \\
\midrule
\rowcolor{lightergray}
Cen-$7$B    & $147.9$ $(1\times)$     & $50.7$ $\boldsymbol{(1\times)}$      & $97.2$ $(1\times)$      & $88$ & $0.105$  \\
\rowcolor{lightgray}
Fed-$7$B    & $95.6$ $\boldsymbol{(0.65\times)}$   & $95.5$ $(1.9\times)$    & $0.1$ $\boldsymbol{(0.001\times)}$   & $90$ & $0.071$  \\
\bottomrule
\end{tabular}
}
\vspace{-0.2cm}
\end{table}

\subsection{\photon Model Performance Scales with Federation Size}\label{sec:performance_scales_w_size}

Previous works \citep{OnLargeCohortTraining,DiLoCo} raised concerns about the wall time benefits of increasing the federated population size.
Using \photon, we show that client contributions to the global model convergence depend on the amount of local work, based on the following hyperparameters: the number of local training steps per round and the global batch size $B_g=NB_l\in\{32,64,128,256,512\}$, where $N\in\{1,2,4,8,16\}$ is the number of clients per round, and $B_l=32$ is the local batch size.

\begin{figure}[t]
    \centering
    \includegraphics[width=\columnwidth]{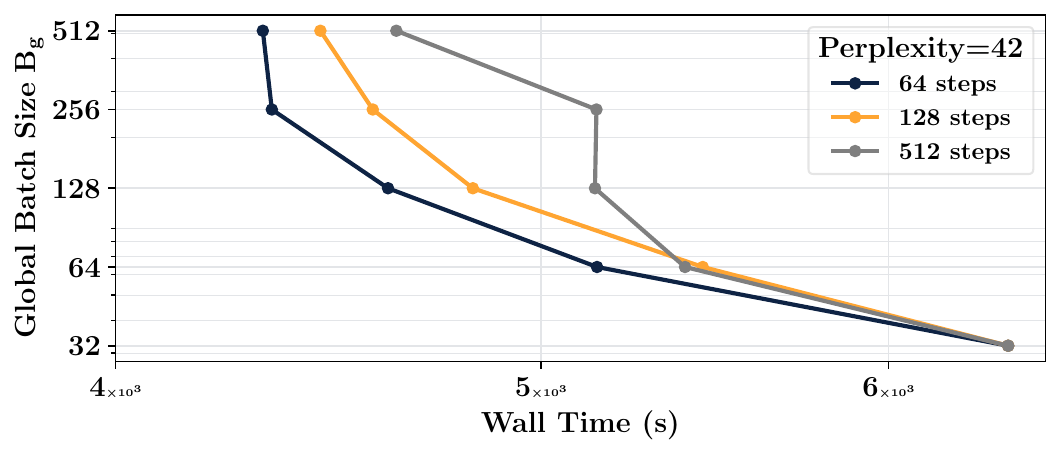}
    \includegraphics[width=\columnwidth]{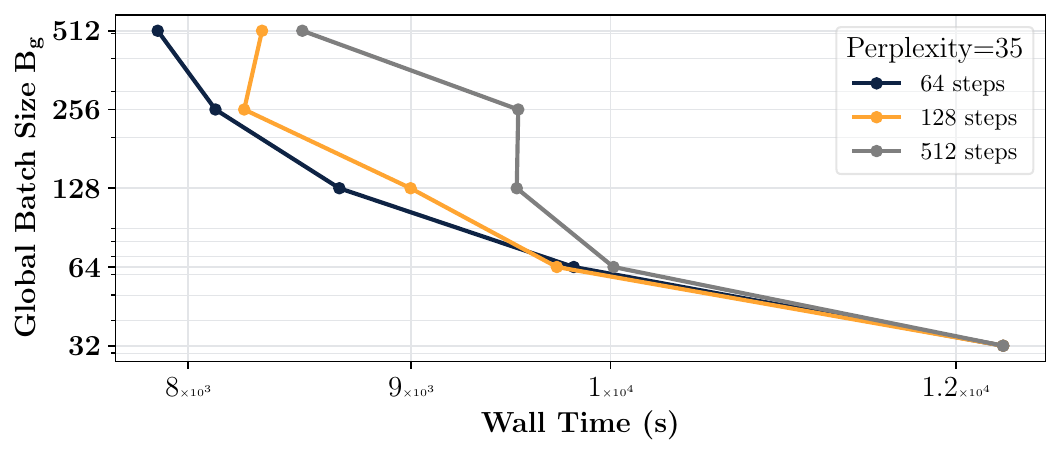}
    \vspace{-0.7cm}
    \caption{\textbf{The tradeoff between time and compute resources (the larger batch size, the more resources) spent to train a model by \photon to target perplexities (top $42$ and bottom $35$).} We measure the impact of the global batch size $B_g = NB_l$, where $N\in\{1,2,4,8,16\}$ (number of clients per round) and $B_l=32$ (local batch size), on the wall time needed to reach two target perplexities: $42$ (top, near the centralized baseline) and $35$ (bottom, near optimum). Fewer local steps per round (64) show clear benefits from increasing $B_g$ for both perplexity targets, but with more local work (128 and 512 steps), the returns on reduced wall time diminish.}
    \label{fig:wall_time_federation_size}
    \vspace{-0.3cm}
\end{figure}

As illustrated in \Cref{fig:wall_time_federation_size}, more frequent communication (smaller local steps per round) leads to greater wall time reductions with larger $B_g$.
The trends are smoother for a higher target perplexity of $42$ than at a lower target perplexity.
On the other hand, at a target perplexity of $35$, the gains in wall time drop as $B_g$ increases, especially at $128$ steps per round. This phenomenon is also observed in \citet{LargeBatchTraining}, where increasing the batch size in the centralized setting led to diminishing returns in wall time. For more details, see \cref{app_subsec:fl_llm_pretrain}. 
Hence, depending on the target perplexity and local steps per round, a compute-optimal regime exists where the global batch size $B_g$ must be carefully tuned to maximize the number of available clients.  \looseness=-1

\textbf{Comparison with State-of-the-Art.} Through utilizing the aforementioned compute-optimal regime, \photon significantly outperforms previous state-of-the-art LLM distributed pre-training framework, DiLoCo~\citep{DiLoCo}. In \Cref{tab:diloco_comparison}, we train a model with $125$M parameters, varying the number of clients $N\in\{2, 4, 8\}$, and using a local batch size $B_l=32$.
We show that, within an appropriate global batch size $B_g$ regime, DiLoCo yields limited returns as the number of clients per round increases and requires roughly $2\times$ more time to reach the same perplexity for both target values, $42$ and $35$.
We observe similar trends across different $B_g$ settings used in DiLoCo.

\begin{figure}[t]
    \centering
    \includegraphics[width=\columnwidth]{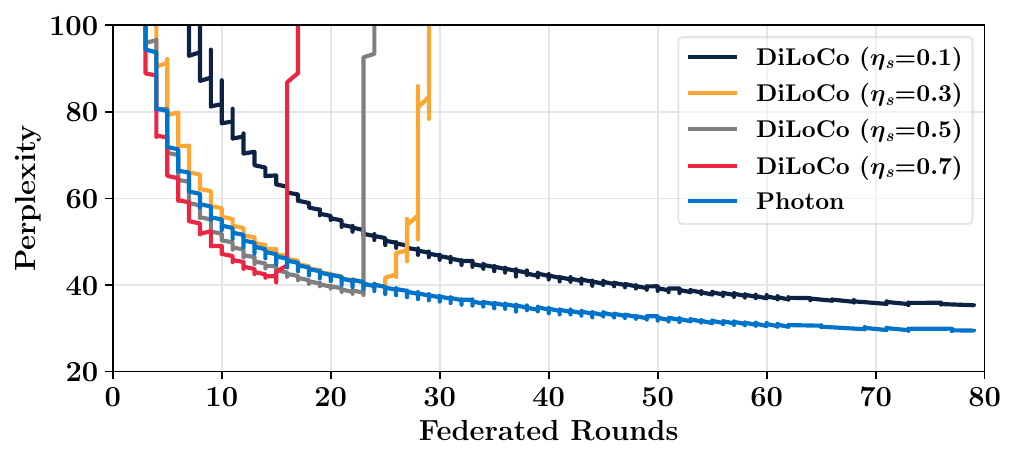}
    \vspace{-0.8cm}
    \caption{\textbf{Perplexity convergence comparison between \photon and DiLoCo.} We tuned DiLoCo's recommended server optimizer (\texttt{OuterOpt}), \texttt{SGD} with Nesterov momentum, using learning rates $\eta_s \in \{0.1, 0.3, 0.5, 0.7\}$, while keeping the momentum coefficient fixed at 0.9. A $125$M-parameter model was trained with a global batch size $B_g = 128$ and $N = 4$ clients per round. Higher $\eta_s$ values accelerated training but hindered achieving the desired perplexity due to early divergent activations. Consequently, we set $\eta_s = 0.1$ in later experiments to reach lower perplexities.}
    \label{fig:diloco_server_lr}
    \vspace{-0.3cm}
\end{figure} 

\begin{table}[t]
\caption{
\textbf{\photon consistently reaches a satisfactory perplexity twice as fast as DiLoCo ($\eta=0.1$) across three client counts.} We evaluated the impact of the global batch size $B_g = NB_l$, where $N \in \{2, 4, 8\}$ (clients per round) and $B_l = 32$ (local batch size), on the wall time required to reach target perplexities of $42$ (near the centralized baseline) and $35$ (near optimal). Results show the wall time gap between \photon and DiLoCo when tuning the server learning rate $\eta_s = 0.1$ to meet the target perplexities.
}\label{tab:diloco_comparison}
\vspace{-0.3cm}
\resizebox{\columnwidth}{!}{
\begin{tabular}{@{}clcc@{}}
\\\toprule
\multirow{2}{*}{$\boldsymbol{N}$}   & \multirow{2}{*}{\textbf{Method}}  & \multicolumn{2}{c}{\textbf{Wall Time [s]}}    \\
                                    &                                   & $\mathbf{PPL}\boldsymbol{=42}$        & $\mathbf{PPL}\boldsymbol{=35}$\\\midrule
\multirow{2}{*}{$2$}                & DiLoCo $(\eta_s=0.1)$             & 10528.8 $(1\times)$                   & 19516.8 $(1\times)$       \\
                                    & \cellcolor{lightergray} \photon   & \cellcolor{lightergray} 5392.8 $\boldsymbol{(0.51\times)}$    & \cellcolor{lightergray} 10015.2 $\boldsymbol{(0.51\times)}$    \\ \midrule
\multirow{2}{*}{$4$}                & DiLoCo $(\eta_s=0.1)$             & 10545.2 $(1\times)$                   & 19032.8 $(1\times)$       \\
                                    & \cellcolor{lightergray} \photon   & \cellcolor{lightergray} 5144.0 $\boldsymbol{(0.49\times)}$    & \cellcolor{lightergray}9516.4 $\boldsymbol{(0.5\times)}$       \\ \midrule
\multirow{2}{*}{$8$}                & DiLoCo $(\eta_s=0.1)$             & 9523.8 $(1\times)$                    & 20334.6 $(1\times)$       \\
                                    & \cellcolor{lightergray} \photon   & \cellcolor{lightergray}5148.0 $\boldsymbol{(0.54\times)}$    & \cellcolor{lightergray}9523.8 $\boldsymbol{(0.47\times)}$     \\ \bottomrule
\end{tabular}
}
\vspace{-0.2cm}
\end{table}

In \citet{DiLoCo}, the authors adopted a much higher compute regime.DiLoCo trains a \textit{smaller} model ($75$M parameter) with $B_l=512$, $N=8$, and $B_g=\num{4096}$ for \num{88000} steps, which corresponds to $46$B tokens per worker and $369$B tokens in total, far beyond the compute-optimal $1.5$B estimated by \citet{TrainingComputeOptimalLLMs}\footnote{The original work is not explicit if $512$ is global or local. Substituting $B_l =\frac{512}{8}$ gives similar conclusions, with $5.75$B tokens per worker and $46$B in total, far beyond the optimal $1.5$B.}. In contrast, for the \textbf{larger} $125M$ model, \photon automatically tunes the local batch size $B_l=32$ to suit the hardware capabilities and achieve \num{9000} cumulative training steps per client when using four clients per round, reaching $2.32$B token processed in total, close to the compute-optimal $2.5$B tokens estimated by \citet{TrainingComputeOptimalLLMs}.

\subsection{\photon's Comms. Efficiency and Scalability}

Communication is an overhead in FL, often scaling poorly as the number of clients grows.
In this section, we evaluate the communication scalability of \photon, considering its nontrivial relationship with the faster convergence speed brought by using more clients per round~(\cref{fig:wall_time_federation_size}).
Thus, we examine three aggregation implementations for \photon: \textit{parameter server} (PS), \textit{AllReduce} (AR), and \textit{Ring-AllReduce} (RAR).
These approaches are detailed in \cref{sec:systemx_implementation}.

Increasing the number of clients per round speeds up convergence, reducing the required communication steps.
Thus, we evaluate communication scalability by varying the number of clients per round, $N\in\{2, 4, 8, 16\}$ (excluding $N=1$, as no communication occurs).
Since the number of pseudo-gradients communicated and aggregated scales linearly with the number of clients and model size, we use the most communication-efficient setup in our experiments.
Specifically, clients perform $512$ local steps per round, training a $125$M parameter model targeting a perplexity of $35$.

\begin{figure}[t]
    \centering
    \includegraphics[width=\columnwidth]{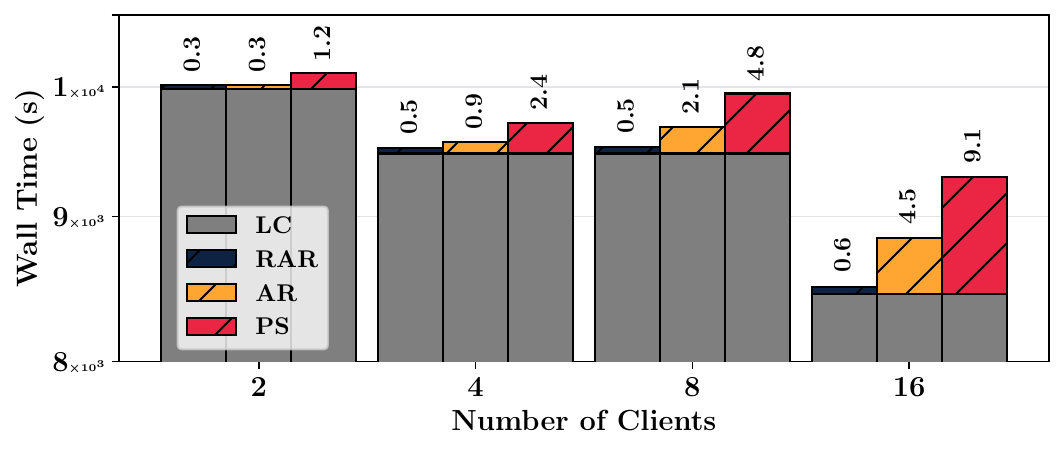}
    \vspace{-0.8cm}
    \caption{\textbf{Wall time comparison among different topologies.} 
    We report total wall time divided into local compute (LC) and communication time. Communication time is evaluated for three aggregation methods: \textit{parameter server} (PS) for privacy-constrained settings, \textit{AllReduce} (AR) for better scalability, and \textit{Ring-AllReduce} (RAR), the most scalable but limited by the slowest link. As expected, communication costs rise with more clients. However, efficient implementations like RAR maintain the wall time reduction from scaling compute resources. The top indicates the percentage of time spent on communication for each method.\looseness=-1}
    \label{fig:wall_time_split_barplot}
    \vspace{-0.4cm}
\end{figure}

\Cref{fig:wall_time_split_barplot} shows that communication overhead increases with $N$, especially under the PS implementation due to its limited scalability. However, using more clients accelerates convergence, reducing the required communication steps and mitigating the costlier communication. Except at the largest cohort size ($16$ clients), the less scalable implementations (PS and AR) account for a minor portion of wall time relative to the local computation (LC) time at clients. \Cref{fig:wall_time_split_barplot} highlights RAR as the most scalable option when minimizing communication time is crucial.

\subsection{\photon's Robustness to Data Heterogeneity}\label{sec:eval_robust}

In typical federated learning scenarios, clients possess heterogeneous data distributions, raising concerns about the robustness of federated optimization. Non-coherent pseudo-gradient updates across clients can impact model convergence by slowing it down or reducing performance. This challenge is compounded by privacy restrictions that prevent observing client data distributions directly.

\begin{figure}[t]
    \centering
    \includegraphics[width=\columnwidth]{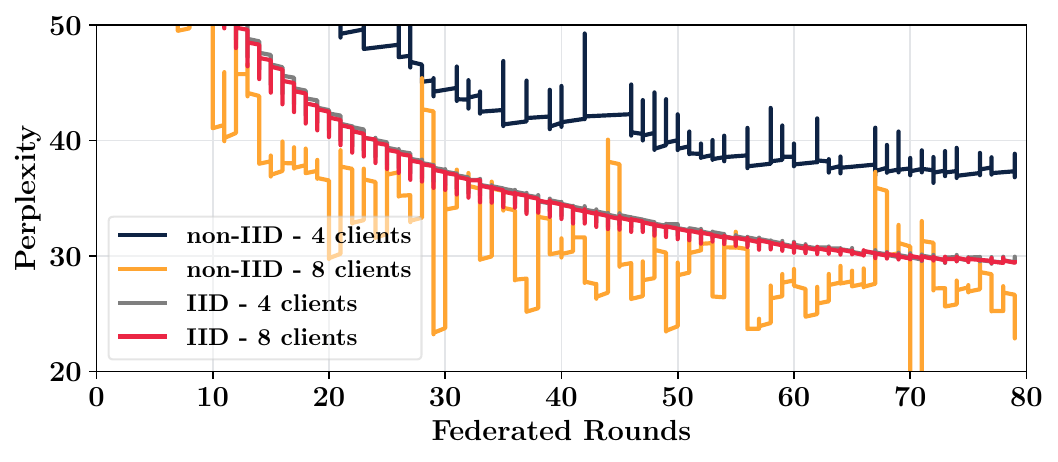}
    \includegraphics[width=\columnwidth]{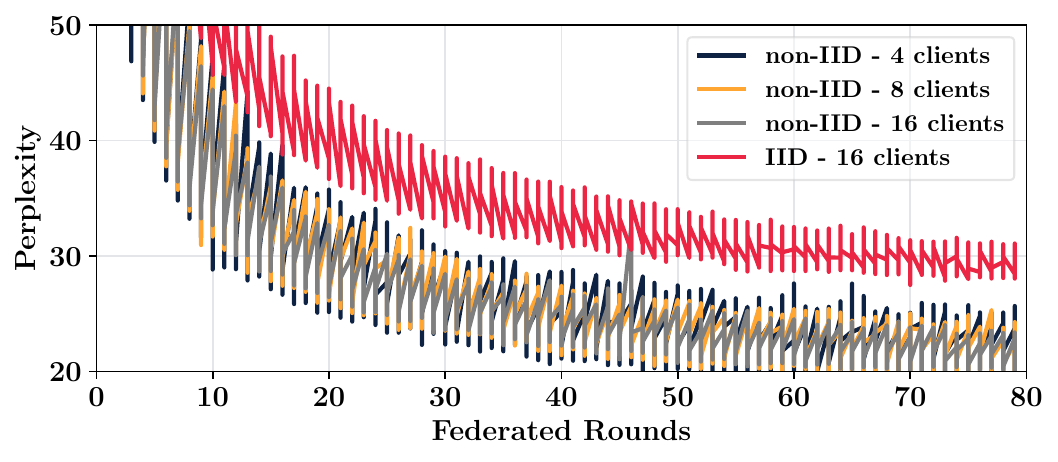}
    \vspace{-0.8cm}
    \caption{\textbf{We assess \photon's robustness to data heterogeneity by training with (top) and without (bottom) partial participation.} Federation of clients using data from The Pile dataset, as described in \cref{sec:exp_setup}. Results from homogeneous data distribution (IID) experiments for reference. While partial participation causes larger fluctuations across rounds, full participation behaves similarly to the IID case. In all settings, increasing the number of clients per round accelerates achieving the target perplexity, with the effect more pronounced under partial participation.}
    \vspace{-0.4cm}
    \label{fig:hetero_data_pp}
\end{figure}

We assess \photon's robustness to data heterogeneity by training on different data sources from The Pile dataset, distributed across clients as described in \cref{sec:exp_setup}. The most challenging scenario involves partial participation of the client population, where the global federated model is only intermittently exposed to diverse data distributions. \Cref{fig:hetero_data_pp} (top) shows that higher client sampling ratios improve convergence speed, final performance, and convergence smoothness under partial participation, demonstrating \photon's robustness to heterogeneous data.

Under full participation, \Cref{fig:hetero_data_pp} (bottom) indicates that larger client cohorts reduce wall-clock training time, though less effectively than with IID data due to conflicting gradients from diverse sources~\citep{OnLargeCohortTraining}. Aggregation methods designed for heterogeneous data, as in~\citep{TIES_ALGO}, could further enhance convergence in such cases.

\section{Opportunities and Limitations}

Our proposed \photon introduces both algorithmic and system optimization techniques to achieve state-of-the-art LLM pre-training across distributed heterogeneous data.
In this section, we reiterate some of our work's limitations and expand on some of our research's potential opportunities and applications.

\noindent\textbf{Federated Hyperparameter Tuning.} In \cref{sec:performance_scales_w_size}, we emphasize the importance of selecting the global batch size to maximize computational resources and performance. A natural extension of our work would be to investigate other critical hyperparameters, such as the learning rate, learning rate scheduler, and their interaction with batch size. \photon’s significant reduction in pre-training costs for LLMs makes it feasible to leverage existing federated hyperparameter optimization algorithms \cite{flora,fedex} to explore optimal global or per-client hyperparameters. \looseness=-1

\noindent\textbf{Addressing Data Heterogeneity.} Our experiments in Section~\ref{sec:eval_robust} demonstrate \photon's robustness in handling data heterogeneity. However, further exploration of alternative aggregation strategies, loss functions, and client selection methods could enhance performance under such conditions. Techniques include minimizing the Euclidean distance or maximizing agreement between global and local models \cite{FedProx,moon}, reducing local model divergence from the global model \cite{feddyn,scaffold}, and measuring client contributions \cite{wang2019measure,wang2020principled}, with client selection based on their value to the global model \cite{cho2020client,huang2021shapley}. Most of these methods can be directly integrated into \photon.

\noindent\textbf{Continual Pre-training \& Personalization.} A key advantage of using \photon for pre-training LLMs is improved model convergence and performance, offering a stronger initialization for continual pre-training or personalization. This aligns with findings from previous studies \cite{nguyen2022begin,chen2023importance}, which highlight the importance of starting with a strong pre-trained model to stabilize federated training and enhance global model performance. A robust pre-trained model also benefits LLM personalization by enabling further fine-tuning or per-client learning strategies \cite{lee2024fedl2p}.

\noindent\textbf{Cross-device Federated Scenarios.} In this work, we demonstrate how \photon effectively addresses the challenges of cross-silo FL. Another common context is cross-device FL, where clients are often lower-compute devices like mobile phones and IoT devices. To handle the diverse system heterogeneity in these scenarios, \photon can be extended with existing methods proven successful in cross-device FL, such as parameter-efficient fine-tuning \cite{sun2022conquering,zhang2023fedpetuning}, quantization \cite{yoon2022bitwidth}, low-rank decomposition \cite{yao2021fedhm}, pruning \cite{caldas2018expanding,jiang2022model}, and early exits \cite{royson2024reefl,kim2023depthfl,ilhan2023scalefl}.

\section{Related Work}

Recent advances in large-scale FL and distributed deep learning focus on improving scalability, efficiency, and privacy while addressing challenges like device heterogeneity and system constraints. Early efforts, such as \cite{ScaleAndSystemDesign}, emphasize system design in FL for mobile devices, addressing resource limitations and connectivity in applications like Google's Gboard. Asynchronous FL systems like PAPAYA \cite{papaya} improve scalability by allowing clients to update models independently, boosting convergence speed and reducing communication overhead. Platforms like FLINT \cite{flint} simulate real-world FL constraints, benefiting large-scale applications like LinkedIn. In distributed deep learning, PyTorch’s Fully Sharded Data Parallel (FSDP) \cite{FSDP_Pytorch} and Distributed Data Parallel (DDP) \cite{PyTorchDistributed} are crucial for scalable training, while frameworks like Horovod \cite{Horovod} use \textit{Ring-AllReduce} to simplify distributed learning. Tools like Dataset Grouper \cite{DatasetGrouper} help create large-scale datasets, and systems like MAST \cite{mast} optimize workload placement across data centers. ZeRO \cite{ZeROplus,ZeroOffload,FSDP_ZeRO} and PETALS \cite{petals_acl,petals_neurips} extend scalability, enabling training and inference for models with trillions of parameters and decentralized deployment, respectively, providing robust solutions for large-scale ML.

In federated or collaborative LLM training, most efforts have focused on fine-tuning~\citep{FedLLMFineTune,ye2024openfedllm} or inference~\citep{petals_acl,petals_neurips}, with pre-training only recently gaining attention~\citep{DiLoCo,sani2024futurelargelanguagemodel,distro,DEPT,LocalAdamw}.
Previously, federated learning primarily centered on cross-device settings, which are unsuited to pre-training's substantial hardware demands.
\citet{DiLoCo}, discussed in the main work, achieves notable communication reductions in a high-compute training regime.
\citet{sani2024futurelargelanguagemodel} introduces a library for LLM pre-training and provides initial results on large models.
\citet{distro} offers early findings on a peer-to-peer AllReduce approach to Federated Learning.
\section{Conclusion}
This work introduces \photon, the first federated system for decentralized end-to-end LLM pre-training in low-bandwidth, globally distributed settings. \photon enables collaborative training of models up to 7B parameters, outperforming centralized training in perplexity. By pooling client resources, \photon accelerates training as compute scales, and in low-bandwidth conditions, it surpasses standard distributed training by increasing throughput and reducing communication costs. This is achieved through adaptive local parallelism, which dynamically switches between distributed algorithms and low-bandwidth Local SGD based on client connectivity. With small local batch sizes and high learning rates, \photon supports an aggressive training strategy, making it the first cost-effective solution to scale LLM pre-training beyond data centers.

\FloatBarrier

\section*{Acknowledgements}
All costs for the computation used for this work was funded by Flower Labs, and the research conducted by a team of researchers from Flower Labs and The University of Cambridge. Support for university-based researchers came from a variety of sources, but in particular, the following funding organizations are acknowledged: the European Research Council (REDIAL), the Royal Academy of Engineering (DANTE), and the Ministry of Education of Romania through the Credit and Scholarship Agency.
We additionally express our immense gratitude to Pedro Porto Buarque de Gusmão, who helped construct the foundations of this project; Javier Fernandez-Marques for supporting us with invaluable insights; and Christopher Irwin and Andrej Jovanovic for the insightful feedback.

\bibliography{references}
\bibliographystyle{mlsys2025}

\clearpage
\newpage
\appendix

\begin{center}
\Large
\textbf{Appendix}
\end{center}

In this appendix, we provide the details omitted in the main paper and more analyses and discussions.

\begin{itemize}
    \item \autoref{app_sec:hypeparams}: Hyperparameters we used for various experiments in our paper, including architectural details and both centralized and federated hyperparameters.
    \item \autoref{app_sec:implem_details}: Implementation details, which include i) full algorithms (pseudo-codes) of the proposed methods (\autoref{app_subsec:full_algo}); ii) implementation of wall time in the paper (\autoref{app_subsec:wall_time}).
    \item \autoref{app_sec:discussions}: Additional discussions that are helpful for the readers to better understand the background, including federated optimization of LLM pre-training (\autoref{app_subsec:fl_llm_pretrain}).
    \item \autoref{app_sec:additional_eval}: Additional evaluations of the systems, e.g., the downstream evaluations.
    \item \autoref{app_subsec:full_algo}: Full algorithms for Distributed Data Parallelism (DDP) and cross-silo federated learning.
    \item \autoref{app_sec:errata_corrige}: Corrections to the MFU values reported in \cref{tab:federation_scalability}.
\end{itemize}

\section{Hyper-parameters}\label{app_sec:hypeparams}

As shown in Table \ref{tab:model_sizes}, we trained models ranging in size from \num{125} million parameters to \num{7} billion for the causal language modeling task.
We used the tokenizer presented in \cite{eleuther_ai_tokenizer} with a vocabulary size of \num{50368}.
The local optimizer the clients use in our experiments is AdamW \citep{AdamW}, while the server optimizer is FedMom \citep{FedMOM}. For all of our non-DiLoCo experiments, we default to FedAvg with server learning rate $1.0$ and server momentum $0.0$.
The hyperparameters we used are reported in Table \ref{tab:fl_hyperaparams}. We chose to train decoder-only models, although our system could train any LLM architecture because they have become the de-facto standard in language modeling and text generation owed to their sample efficiency. 

We also note that the billion-scale experiments assume intermittent client availability, reflecting real-world scenarios in which participants may occasionally allocate free computing resources to federated pre-training.
To accommodate this, we employ a stateless local optimization procedure, and resetting optimizer states each round. This enables \photon to operate seamlessly in sparse-compute scenarios, unlike standard distributed data parallelism (DDP), which requires fully dedicated and synchronized GPU workers.
Stateless local optimization also eliminates the communication costs of synchronizing optimizer states, making it easier to ensure that federated pre-training remains compute-bound.

\begin{table}[ht]
\centering
\caption{Architecture details and local training parameters for our $125$M, $350$M, $1.3$B, $3$B, and $7$B models. We report the number of transformer blocks, hidden model dimension $d$, number of attention heads, the linear layer expansion ratio, and Adam's parameters ($\beta_1$ and $\beta_2$). We also report the vocabulary size of the tokenizer we used~\citep{eleuther_ai_tokenizer} and the sequence length $l$.}
\label{tab:model_sizes}
\resizebox{\columnwidth}{!}{
\begin{tabular}{@{}rccccccc@{}}
\toprule
    \textbf{Model} &
    \multirow{2}{*}{\textbf{\#Blocks}} &
    \multirow{2}{*}{$\boldsymbol{d}$} &
    \multirow{2}{*}{\textbf{\#Heads}} &
    \multirow{2}{*}{\textbf{Exp.~Ratio}} &
    \multirow{2}{*}{$\mathbf{(\boldsymbol{\beta_1},~\boldsymbol{\beta_2})}$} &
    \multirow{2}{*}{$\boldsymbol{|\textbf{Vocab}|}$} &
    \multirow{2}{*}{$\boldsymbol{l}$} \\
    \textbf{Size} &
    &
    &
    &
    &
    &
    &
    \\
\midrule
    \textbf{$\mathbf{75}$M} & 3 & 896 & 16 & 4  & $(0.9,~0.95)$ & \num{50368} & \num{1024} \\
    \textbf{$\mathbf{125}$M} & 12 & 768 & 12 & 4  & $(0.9,~0.95)$ & \num{50368} & \num{2048} \\
    \textbf{$\mathbf{350}$M} & 24 & 1024 & 16 & 4  & $(0.9,~0.95)$ & \num{50368} & \num{2048} \\
    \textbf{$\mathbf{1.3}$B} & 24 & 2048 & 16 & 4  & $(0.9,~0.95)$ & \num{50368} & \num{2048} \\
\midrule
    \textbf{$\mathbf{3}$B} & 32 & 2560 & 20 & 4  & $(0.9,~0.95)$ & \num{50368} & \num{2048} \\
    \textbf{$\mathbf{7}$B} & 32 & 4096 & 32 & 4  & $(0.9,~0.95)$ & \num{50368} & \num{2048} \\
\bottomrule
\end{tabular}
}
\end{table}

\begin{table}[ht]
\centering
\caption{Hyperparameters used in our experiments. The federated learning rate $\boldsymbol{\eta_s}$ and momentum $\boldsymbol{\mu_s}$~\citep{FedMOM} are applied by the \photon Agg. $\mathbf{S_C}$ are the parameters of the learning rate scheduler synchronized across \textbf{sequential} steps. $\alpha$ is the factor to be applied to the maximum learning rate $\eta_{max}$ to obtain the minimum learning rate for the cosine scheduler, i.e., $\eta_{min}=\alpha\times\eta_{max}$. $T$ is the duration, in steps, of the cosine scheduler for fed/cent variants. We also report the batch size used in the local training by the \photon clients and the centralized batch size.}
\label{tab:fl_hyperaparams}
\resizebox{\columnwidth}{!}{
\begin{tabular}{@{}lcccccccc@{}}
\toprule
\textbf{Model} & \multirow{2}{*}{$\boldsymbol{\eta_s}$} & \multirow{2}{*}{$\boldsymbol{\mu_s}$} & \multirow{2}{*}{$\boldsymbol{\alpha}$} & \multirow{2}{*}{$\boldsymbol{\eta_{max}}$} & \multirow{2}{*}{$\boldsymbol{T}$} & \multirow{2}{*}{$\boldsymbol{T}_{\mathrm{cent}}$} & \textbf{Batch} & \textbf{Batch} \\
\textbf{Size} &  &  &  &  &  &  & \textbf{Size} & \textbf{\begin{tabular}[c]{@{}c@{}}Size\\ Cent\end{tabular}} \\ \midrule
\textbf{$\mathbf{125}$M} & $\{0.0,0.1,0.3 ,0.5,0.7, 1.0\}$ & $\{0.9, 0.0\}$ & $10^{-1}$ & $ 6.0 \times 10^{-4}$ & \num{40960} & \num{5120} & \num{32} & \num{256} \\
\textbf{$\mathbf{1.3}$B} & $1.0$ & $0.0$ & $10^{-1}$ & $2\times10^{-4}$ & \num{24800} & \num{24800} & \num{512} & \num{512} \\
\textbf{$\mathbf{3}$B} & $1.0$ & $0.0$ & $10^{-1}$ & $1.6\times10^{-4}$ & \num{51500} & \num{51500} & \num{512} & \num{512} \\
\textbf{$\mathbf{7}$B} & $1.0$ & $0.0$ & $10^{-1}$ & $1.2\times10^{-4}$ & \num{63900} & \num{63900} & \num{1024} & \num{1024} \\ \bottomrule
\end{tabular}
}
\end{table}

\begin{table}[ht]
\centering
\caption{Hyperparameters for our federated experiments. $P$ represents the total number of clients per federations, $K$ the number of clients sampled per round, $D$ the dataset, $\tau$ the number of steps per round.}
\label{tab:fl_settings}
\resizebox{\columnwidth}{!}{
\begin{tabular}{@{}rcccc@{}}
\toprule
\textbf{Model} & \multirow{2}{*}{$\boldsymbol{P}$} & \multirow{2}{*}{$\boldsymbol{K}$} & \multirow{2}{*}{$\boldsymbol{D}$} & \multirow{2}{*}{$\boldsymbol{\tau}$} \\
\textbf{Size} &  &  &  &  \\ \midrule
\textbf{$\mathbf{125}$M} & $\{1,2, 4, 8, 16\}$ & $\{1, 2,4,8,16\}$ & C4~\citep{C4}, The Pile~\citep{ThePile} & ${64, 128, 512}$ \\
\textbf{$\mathbf{1.3}$B} & $8$ & $8$ & C4~\citep{C4} & $500$ \\ \midrule
\textbf{$\mathbf{3}$B} & $4$ & $4$ & C4~\citep{C4} & $500$ \\
\textbf{$\mathbf{7}$B} & $4$ & $4$ & C4~\citep{C4} & $500$ \\ \bottomrule
\end{tabular}
}
\end{table}
\section{Implementation details} \label{app_sec:implem_details}
\subsection{Modeling Wall Time} \label{app_subsec:wall_time}

We implement a comprehensive wall time model to analyze the temporal efficiency of our federated learning system across different communication architectures.
The wall time calculations account for both computation and communication costs, considering factors such as local training time, model broadcast time, gradient collection time, and aggregation overhead.

\paragraph{Local Training Time}
The local training time ($T_L$) for each client is determined by the number of local training steps and the client's computational throughput:

\begin{equation}
    T_L = \frac{\tau}{\nu},
    \label{eq:local_time}
\end{equation}

where $\tau$ represents the number of local training steps and $\nu$ is the local throughput measured in batches per second.
Notably, $T_L$ doesn't scale with the number of clients per round $K$ as we assume the ideal case where they all execute the same local training recipe in parallel on equipollent hardware.
In our experiments, $\tau$ represents a hyperparameter that we vary to observe its influence on the final performance.
During deployment, $\tau$ is one of the most important hyperparameters to tune to achieve a pre-defined objective, i.e., target perplexity value at some target wall time.
The value of $\nu$ depends on the computing resources available and the distributed strategy that \photon adopts at local clients.
Throughout our evaluation of the $125$M parameter model, we used an empirical value of $\nu = 2$ batches per second for both centralized and federated models. For the $1$B model, we used an empirical $\nu$ value of $0.147$ for federated models and $0.839$ for centralized models. For the $3$B model, we used $\nu = 0.144$ and $0.395$ respectively, and for the $7$B model, we used $\nu = 0.032$ and $\nu = 0.12$.

\paragraph{Communication Time}
The communication overhead varies based on the chosen architecture.
We implement a bandwidth scaling factor for systems with more than $\theta$ channels (default: 100) to account for network congestion.

For Parameter Server (PS) architecture, the total communication time ($T_C^P$) is:

\begin{equation}
    T_C^{PS} = \begin{cases}
        \frac{KS}{B} & \text{if } K \leq \theta; \\[2ex]
        \frac{KS}{B} & \text{if } K > \theta,
    \end{cases}
    \label{eq:ps_comm}
\end{equation}

where:
\begin{itemize}
    \item $K$ is the number of clients per round;
    \item $S$ is the model size in megabytes;
    \item $B$ is the server bandwidth in MBps.
\end{itemize}

For AllReduce (AR) architecture, the communication time ($T_C^{AR}$) is:

\begin{equation}
    T_C^{AR} = \frac{(K-1)S}{B},
    \label{eq:ar_comm}
\end{equation}

For Ring AllReduce (RAR), we optimize the communication pattern, resulting in:

\begin{equation}
    T_C^{RAR} = \frac{2S(K-1)}{KB}.
    \label{eq:rar_comm}
\end{equation}

We admit that accounting for congestion and real-world measurements could further improve these models. However, we find them to provide sufficiently accurate results.

\paragraph{Total Wall Time}
The total wall time for one round ($T_r$) combines local computation and communication costs:

\begin{equation}
    T_r^\alpha = T_L + T_C^\alpha,
    \label{eq:round_time}
\end{equation}

where $\alpha \in \{PS,AR,RAR\}$ represents the chosen architecture.

The total wall time for the entire training process ($T$) is:

\begin{equation}
    T^\alpha = RT_r^\alpha,
    \label{eq:total_time}
\end{equation}

where $R$ is the total number of federated learning rounds.

The aggregation time $T_{agg}$ is calculated by:
\begin{equation}
    T_{agg} = \frac{KS}{\zeta},
    \label{eq:agg_time}
\end{equation}
where $\zeta$ is the server computational capacity.
The default value of $\zeta$ is 5TFLOPS per second.
For simplicity, the aggregation time at the server is currently considered negligible compared to communication costs, as shown in Equation~\ref{eq:total_time}.
Still, the model allows for future extensions to include server-side computational overhead in cases where its computational capabilities are highly constrained.
Our implementation accounts for exceptional cases as well, such as single-client scenarios without communication.

\subsection{Performance impact of the \textit{Link} component}

The \textit{Link} component provides the crucial connection between the aggregator~(Agg) and client~(LLM-C).
The bandwidth available to the link of each LLM-C dictates how quickly the model parameters can be exchanged for aggregation.
As discussed in \cref{app_subsec:wall_time}, in standard Distributed Data Parallel~(DDP) training, gradients have to be synchronized before every gradient descent step.
The most efficient implementations use the \textit{Ring-AllReduce} algorithm to synchronize gradients while having workers communicate in a peer-to-peer fashion using RDMA network, such as \texttt{InfiniBand} and \texttt{RoCE}.
When using the same aggregation/synchronization algorithm, e.g., \textit{Ring-AllReduce}, for both DDP and \photon, the \textit{Link} bandwidth determines the gap in communication time between the two methodologies.
Using the methodology presented in \cref{app_subsec:wall_time} and assuming a given aggregation algorithm, one can estimate the communication time for a given model size and bandwidth.
Factoring in that \photon communicates less frequently by a constant factor \(\tau\) (e.g., \(500\times\)), the minimal \textit{Link} bandwidth \(B_{\mathrm{Photon}}\) required for \photon to match DDP’s communication time at bandwidth \(B_{\mathrm{DDP}}\) follows from:
\[
    B_{\mathrm{Photon}} \;\ge\; \frac{B_{\mathrm{DDP}}}{\tau}.
\]

This ignores the optimization aspects of model training, which may increase the compute time of \photon, which is why it represents a mere minimum bandwidth.
If the above inequality does not hold, it is necessary to either increase the available bandwidth or to make the communication of \photon more infrequent, which may impact the machine learning performance.

\subsection{Overlapping communication and cleaning up}

When partial participation is involved in the federated setting, clients may sporadically become available or drop out of the federation at any time.
When they disconnect after they finish executing their local work for a specific federated round, the \photon LLM Client can offload the communication process and simultaneously clean up the memory allocated by the training pipeline to allow for the prompt return to idle state.
This routine quickly makes computing resources available for another workload, which is particularly important when using shared computing facilities. 

\subsection{Advanced sharing for reducing memory footprint}

Every \photon LLM Client comprises a multiprocessing stack managed by a leader process that coordinates subordinate processes handling the hardware accelerators.
Such a leader process is also in charge of the communication endpoint, so it receives and sends model parameters as the algorithm requires.
To minimize the RAM footprint up to $8\times$, the model parameters exchanged are stored in shared memory, accessible by all subordinate processes.
\section{Extended Discussion} \label{app_sec:discussions}
\subsection{Federated Optimization of LLM Pre-training} \label{app_subsec:fl_llm_pretrain}

Federated optimization differs significantly from standard data-parallel training due to infrequent synchronization, which affects the multitude of assumptions upon which centralized pre-training of LLMs is built. In a centralized context, previous works have shown the following: (a) the number of parameters $|\theta|$ and the number of tokens $T$ seen by the model should be scaled roughly equally~\citep{TrainingComputeOptimalLLMs} for compute-optimal training; (b) the batch size $\mathcal{B}$ should be chosen based on the available hardware resources, with larger batch sizes providing benefits until a critical batch size $\mathcal{B}_{\mathrm{crit}}$, is reached~\citep{LargeBatchTraining}; (c) the learning rate should be scheduled using cosine decay with a period equal to the total number of optimization-steps/batches $T/\mathcal{B}$. All of these components need to be modified for effective federated optimization.  

From a theoretical perspective, infrequent parameter averaging methods such as Local SGD~\citep{DontUseLargeBatchesUseLocalSGD} or FedAvg~\citep{fedavg} are expected to provide an effect similar to scaling the batch size in a centralized setting~\citep{LinearSpeedupCommsEfficientSGD} when scaling the number of clients per round, however, given the many moving parts of the centralized recipe obtaining such improvements requires successfully adapting it to a federated context. We need to distinguish between the batch size of a given client $\mathcal{B}_{c}$ and the effective batch size of a given round $\mathcal{B}_{\mathrm{eff}} = \sum_{c \in C_r} B_c$, which depends on the batch size of all sampled clients. While smaller batch sizes are known to provide generalization benefits~\citep{SharpMinimaLargeBatch}, for the sake of efficiency, using the largest batch size that can fit inside a given accelerator is preferable. Thus, we assume that each client uses a fixed $B_c$ determined by their hardware and that, for the sake of simplicity, all clients have access to the same hardware \[\mathcal{B}_{i} = B_{j}\,, \forall i,j \in C \times C.\]
In cases where clients have sufficiently powerful hardware, we assume that they use a batch size $\mathcal{B}_c = \mathcal{B}_{\mathrm{crit}}$ to avoid wasting compute. With this simplifying assumption, the compute-time trade-off in federated optimization depends only on the number of clients sampled in a given round $|C_r|$ and the number of local iterations $T_{c}$ performed on each client, both assumed constant. In the case of $T_{c} = 1$, this coincides, when assuming full participation, with the centralized setting, allowing the critical batch size $\mathcal{B}_{\mathrm{crit}}$ to be determined using the gradient noise scale as done in the work of \citet{LargeBatchTraining}.  We perform numerous experiments to understand how the number of local steps $T_{c}$ changes the Pareto-frontier of the compute-time trade-off.

Assuming that the findings of \citet{TrainingComputeOptimalLLMs} hold in a federated context, the compute-optimal number of total steps, $T = R \times T_{c}$, that should be performed depends on the number of tokens appropriate for the given model size, roughly $20 \times |\theta|$ according to the work of \citet{TrainingComputeOptimalLLMs}, and on the effective batch size as follows:

\begin{align}
    R \times T_{c} = \frac{20 \times |\theta|}{\mathcal{B}_{\mathrm{eff}}},
    \label{eq:optimal_num_steps}
\end{align}

with the large caveat that this compute-optimal point was chosen, assuming that training would be conducted using the centralized critical batch size.
However, accounting for this in the learning rate schedule is not trivial for two reasons. First, the averaging-based aggregation procedure will limit the impact of any individual update. This is true both from a simple mathematical perspective, the norm of the average update being less than the average of the update norms, and because client updates in federated learning have been shown to produce near-orthogonal updates which tend to result in small pseudo-gradient norms~\citep{OnLargeCohortTraining}.  Second, clients take several optimization steps using their hardware-determined batch size before aggregation, with smaller batch sizes being known to require smaller learning rates~\citep{LargeBatchTraining} in centralized settings. Since we expect the client hardware batch size to be generally smaller than $\mathcal{B}_{\mathrm{crit}}$, they are likely to fall in the regime of small-batch training. Thus, in our work, we propose decaying the learning rate following a cosine scheduler appropriate for the hardware batch size $B_c$, using \cref{eq:optimal_num_steps} to obtain the period by replacing $\mathcal{B}_{\mathrm{eff}}$ with $\mathcal{B}_{c}$. Having fixed this period, we only have to tune one hyperparameter in the form of the maximum learning rate $\eta_{\max}$ with the minimum learning rate being computed as $\eta_{\min} = \frac{\eta_{\max}}{10} $. In contrast, we find that neither square root nor linear learning rate scaling~\citep{LRScaling,SurgePhenomenaLR} sufficiently stabilize centralized training across varying batch sizes. 

The momenta stabilize the local optimization direction for a given local momentum-based optimizer, such as AdamW~\citep{AdamW}. Since they are generally implemented using exponential decay, the impact of any individual gradient step is reduced. In the case of federated optimization, this poses a challenge as we aim to update the momentum vectors to reflect the information received during the aggregation step. Directly communicating and averaging the momenta of all clients would increase the communication costs by the number of momenta of the optimizer relative to transmitting only the parameters, e.g., it would be three times higher for AdamW. To avoid such increases in communication costs, we keep the momenta local and rely on only the parameter update to regularize training.

\subsection{Recent Advances in Federated Optimization}

Recent works such as \citet{LocalAdamw} and \citet{DEPT} have shown that federated optimization algorithms can be competitive with standard Distributed Data-Parallel methods in specific circumstances.

\citet{LocalAdamw} prove the convergence of a Local AdamW variant that averages both model parameters and optimizer states every round.
They further show that, depending on the task and data distribution, local-update optimizers can converge faster than standard minibatch SGD, particularly under IID data across workers/clients (where update variance is low).

\citet{DEPT} demonstrate the effectiveness of federated optimization even when data heterogeneity is high.
They observe that the noise introduced by averaging model updates from diverse data distributions can yield a more robust set of parameters for the transformer body, potentially improving generalization or adaptation to new data distributions.
\section{Additional Evaluation} \label{app_sec:additional_eval}

\begin{figure}[ht]
    \centering
    \includegraphics[width=\columnwidth]{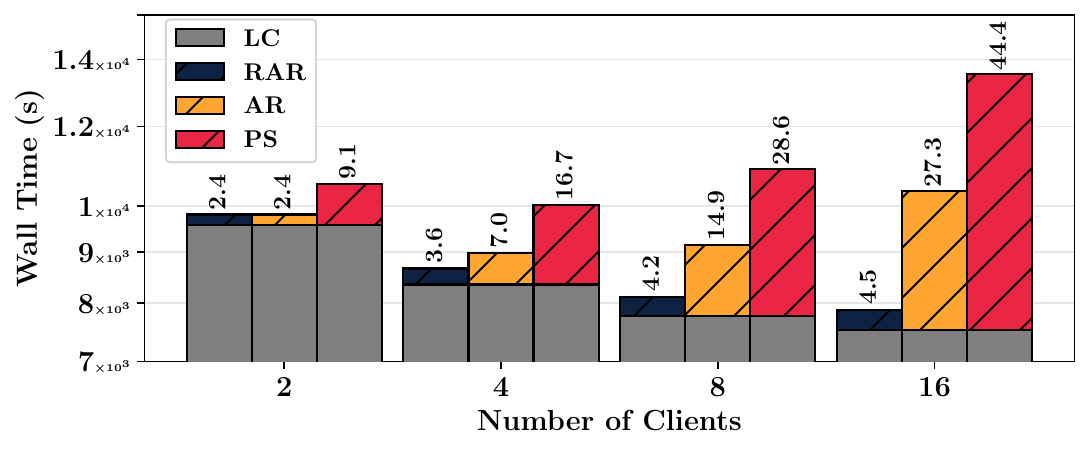}
    \caption{We report the split of the total wall time in two parts: the local compute time (LC) the clients endure to achieve the desired perplexity value and the communication time. The communication time is reported for three different aggregation implementations: \textit{parameter server} (PS), which is necessary when privacy constraints are present; \textit{AllReduce} (AR), more scalable than PS; \textit{Ring-AllReduce} (RAR), the most scalable approach bounded by the slowest link across the ring topology. As we expected, communication represents a more important cost as the number of clients increases. Still, when implemented efficiently (RAR), the wall time benefits of scaling the computing resources are maintained. At the top of each bar, we report the percentage of time spent communicating for the respective experiments and implementation.}
    \label{fig:wall_time_split_64steps_barplot}
\end{figure} 

\subsection{Communication Efficiency and
Scalability}

\begin{figure}[ht]
    \centering
    \includegraphics[width=\columnwidth]{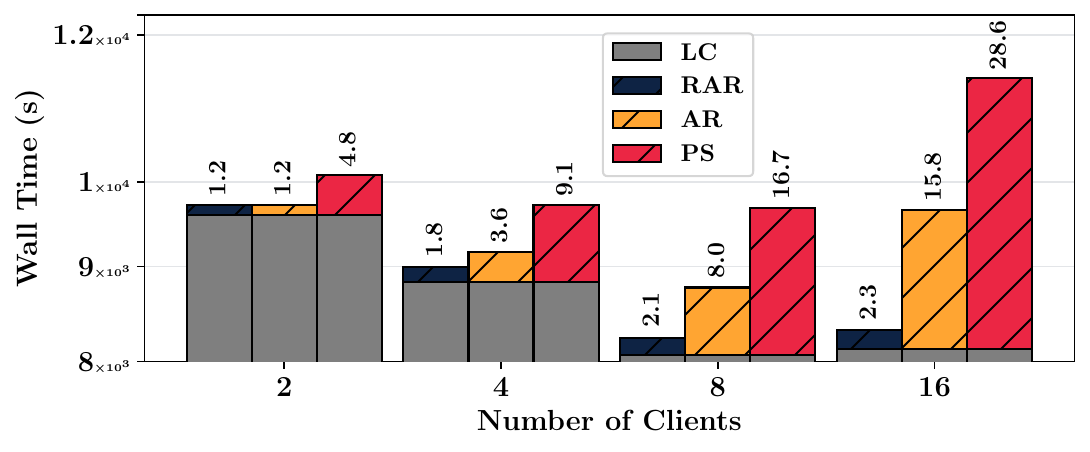}
    \caption{We report the split of the total wall time in two parts: the local compute time (LC) the clients endure to achieve the desired perplexity value and the communication time. The communication time is reported for three different aggregation implementations: \textit{parameter server} (PS), which is necessary when privacy constraints are present; \textit{AllReduce} (AR), more scalable than PS; \textit{Ring-AllReduce} (RAR), the most scalable approach bounded by the slowest link across the ring topology. As we expected, communication represents a more important cost as the number of clients increases. Still, when implemented efficiently (RAR), the wall time benefits of scaling the computing resources are maintained. At the top of each bar, we report the percentage of time spent communicating for the respective experiments and implementation.}
    \label{fig:wall_time_split_128steps_barplot}
\end{figure}

We present additional results on the communication efficiency and scalability of \photon when using 64 local steps per round(~\cref{fig:wall_time_split_64steps_barplot}) and 128 local steps per round(~\cref{fig:wall_time_split_128steps_barplot}). While using 128 steps results in a slightly increased total computational load due to minor reductions in convergence speed, reducing communication frequency by half significantly lowers the communication burden, particularly with higher numbers of clients per round. This trend is especially pronounced in communication-inefficient PS implementations and also applies to the faster RAR and AR methods.

\subsection{\photon Robustness to Node Failures}\label{app_sec:robustness_node}

In centralized data center training, strong synchronization and a fixed communication topology mean that a single hardware failure can halt training, requiring a complete restart from a past checkpoint.
Such hardware failures, even limited to one accelerator, are common and account for $98\%$ of training restarts \citep{llama3}.
In contrast, our federated approach offers a more robust and communication-efficient alternative to distributed data parallelism (DDP).
\photon only needs one pseudo-gradient update to progress a federated round while asynchronously restarting edge components (LLM Clients), unlike centralized systems that require a full restart to reinitialize the distributed process group \citep{PyTorchDistributed}.
Thus, \photon is completely robust to any failure affecting less than $100\%$ of the system.
The decoupling between LLM Clients and the data source, along with the Aggregator's ability to seamlessly add new Clients, allows \photon to maintain the same process group even as Clients are added or removed.

\begin{figure}[ht]
    \centering
    \includegraphics[width=\columnwidth]{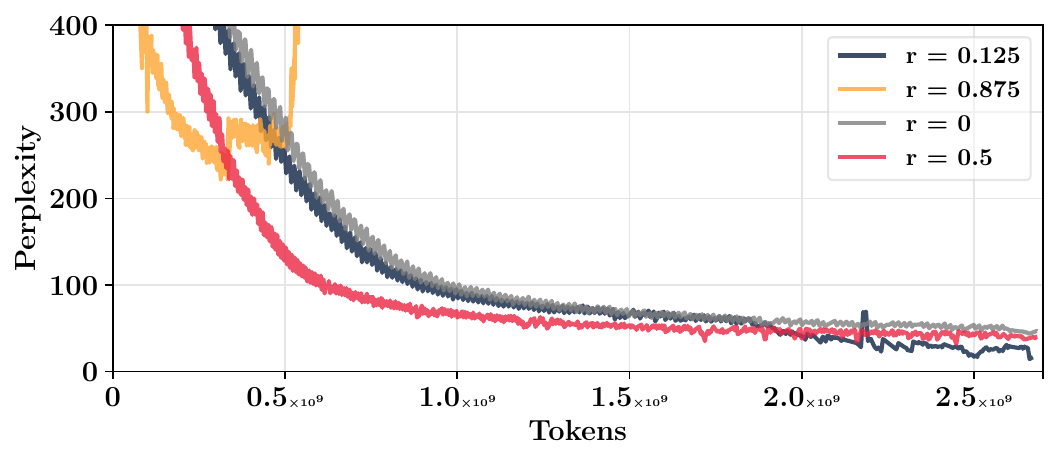}\\
    \includegraphics[width=\columnwidth]{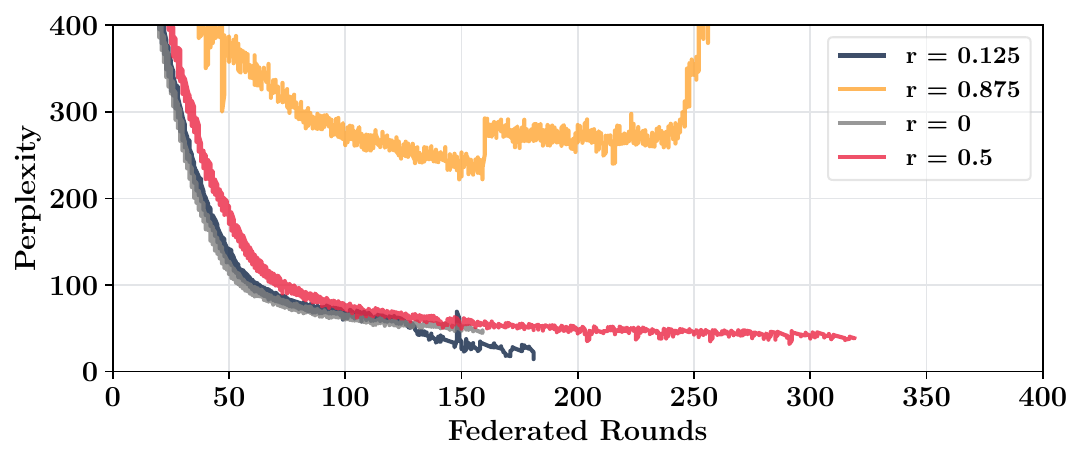}
    \caption{The robustness of \photon for IID data distributions across clients. We show the training perplexity against (top) the number of tokens trained and (bottom) the number of federated rounds for different dropout ratios $r\in\{0, 0.125, 0.5, 0.875\}$ corresponding to $\{0, 1, 4, 7\}$ clients dropping out at every federated round respectively. With the highest value of $r$, the training procedure fails to converge as there is not sufficient training data per round to leverage the hyperparameter setting. For all other values of $r$, the federated training succeeds, potentially reaching a final better perplexity with the same number of total tokens. However, values of $r>0$ result in longer training executions as the number of federated rounds to reach the target number of total tokens increases proportionally to the number of clients dropping out.}
    \label{fig:robustness_8_iid}
\end{figure} 

To test the robustness of \photon against node failures, we run a series of experiments simulating various node dropout ratios.
We configured the federated pre-training of a $125$M parameters model as if we were not expecting any failure between our clients: $8$ clients per round, $32$ samples in each local batch, $32$ local steps per round, and a target number of total tokens to train on equal to $\sim2.5\times10^9$ ($5120$ sequential steps and $160$ federated rounds at full client participation), i.e., $20$ token per parameter considering that we used a model with $125$M parameters.
The other training hyperparameters were the same as the main paper experiments referring to the $125$M parameter models unless stated otherwise.

This setting corresponds to the centralized environment where at least a GPU in any node fails every $32$ training steps, for a total of $N_{\text{failures}} = \frac{\num{5120}}{32} = 160$, i.e., a rate of $3.125\%$ failures per step.
Notably, for the standard centralized approach, different numbers of GPUs or nodes failing may have the same impact on the training procedure as the entire process group often needs to be restarted.

\begin{figure}[ht]
    \centering
    \includegraphics[width=\columnwidth]{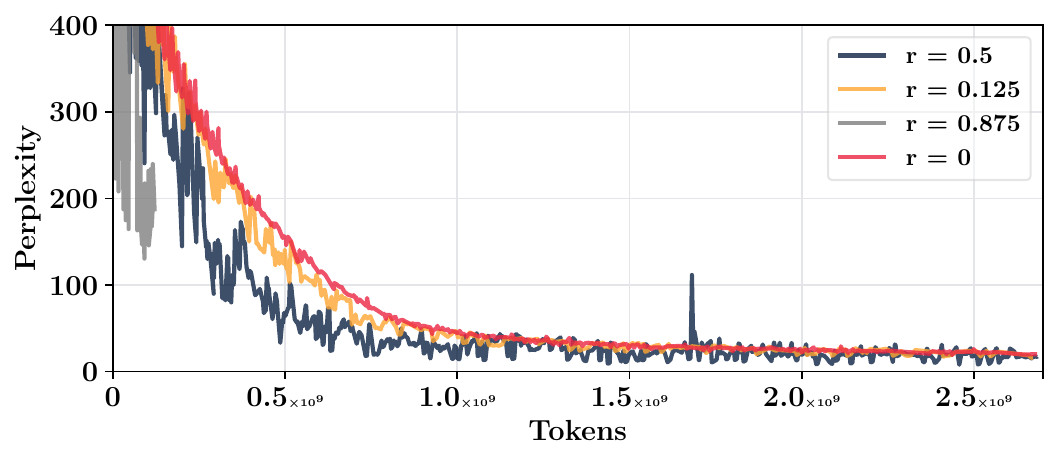}\\
    \includegraphics[width=\columnwidth]{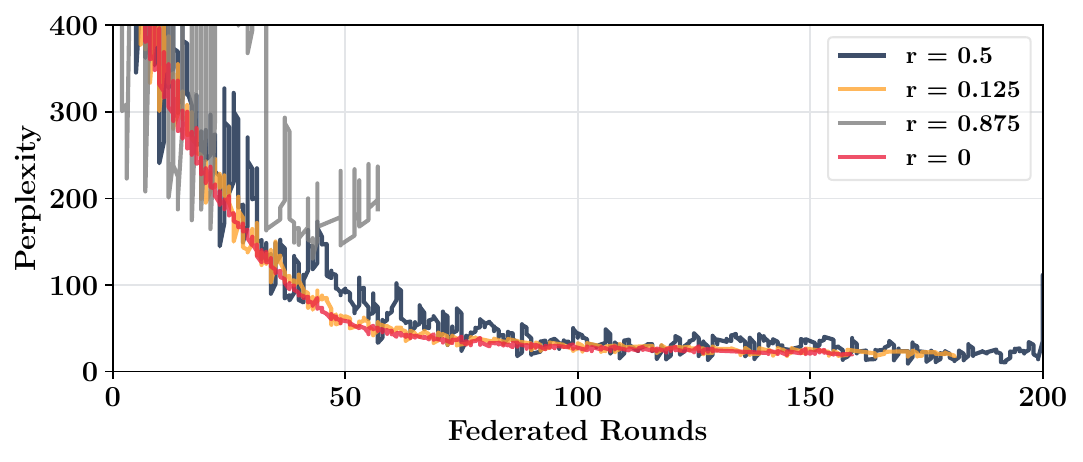}
    \caption{The robustness of \photon for non-IID data distributions across clients. We show the training perplexity against (top) the number of tokens trained and (bottom) the number of federated rounds for different dropout ratios $r\in\{0, 0.125, 0.5, 0.875\}$ corresponding to $\{0, 1, 4, 7\}$ clients dropping out at every federated round respectively. With the highest value of $r$, the training procedure fails to converge as there is not sufficient training data per round to leverage the hyperparameter setting. For all other values of $r$, the federated training succeeds, potentially reaching a final better perplexity with the same number of total tokens. However, values of $r>0$ result in longer training executions as the number of federated rounds to reach the target number of total tokens increases proportionally to the number of clients dropping out.}
    \label{fig:robustness_8_noniid}
\end{figure}

We model IID data distribution across clients by randomly partitioning the \texttt{C4} \citep{C4} dataset uniformly into $8$ equally sized shards.
For nonIID experiments, unlike the main work, which uses the well-known \texttt{C4} dataset, we adopt the newer state-of-the-art data mixture used by \texttt{SmolLM-V2}~\citep{SmolLMV2}, specifically we randomly partition each of the following datasets into $16$ IID shards: (1) \texttt{FineWeb-EDU}~\citep{FineWeb}, a high-quality general language dataset sourced from \texttt{Common Crawl}, (2) \texttt{Cosmopedia V2}~\citep{Cosmopedia}, a synthetic dataset generated by the \texttt{Mixtral-8x7B-Instruct-v0.1} model, (3) \texttt{Python-EDU}, a high-quality subset of \texttt{The Stack V2}~\citep{Stack2} code dataset, (4) \texttt{FineMath 4+}~\citep{SmolLMV2}, a high-quality math subset of  \texttt{Common Crawl}, (5) \texttt{Infi-WebMath 4+}, a high-quality variant \texttt{Infi-WebMath}~\citep{InfiniWebMath} released by \citet{SmolLMV2}. Following the recipe of \citet{SmolLMV2}, we then compose shards to construct clients whose data is comprised of: $70\%$ \texttt{FineWeb-EDU} data, $10\%$ \texttt{Cosmopedia V2}, $10\%$ \texttt{Python-EDU}, $5\%$ \texttt{FineMath 4+}, and $5\%$ \texttt{Infi-WebMath 4+}.

The relevant comparisons we show in this evaluation relate to how the convergence, in terms of local training perplexity, is impacted by the absence of updates due to clients dropping out.
\Cref{fig:robustness_8_iid,fig:robustness_8_noniid} show that only extreme and unrealistic dropout ratios ($r=0.875$) can completely disrupt the training independently on the heterogeneity of the data distributions.
Notably, for the other values of $r$ and for both IID and non-IID data, more dropouts correspond to better final perplexity when effectively training on the same number of total tokens, i.e., executing far more federated rounds (taking more time).
When comparing the final perplexity at different values of $r$ with the number of federated rounds, which are directly proportional to the real wall time, more clients dropping out result in longer training times to achieve the target number of total tokens, as expected.

\begin{table}[t]
\caption{In-context learning comparison between \photon models. Our biggest model wins $3$ out of $3$ comparisons in this group.}
\resizebox{\columnwidth}{!}{
\begin{tabular}{@{}lcccc@{}}
\toprule

\textbf{Name}
& \begin{tabular}[c]{@{}c@{}}\textbf{ARC-Challenge}\\ \citep{arc_challenge}\end{tabular}
& \textbf{\begin{tabular}[c]{@{}c@{}}BigBench \\ QA Wikidata\\ \citep{bigbench}\end{tabular}}
& \begin{tabular}[c]{@{}c@{}}\textbf{HellaSwag}\\  \citep{hellaswag}\end{tabular} \\
\midrule
\rowcolor{lightgray} \textbf{\photon-7B}
& $\mathbf{0.265}$
& $\mathbf{0.447}$
& $\mathbf{0.524}$\\

\textbf{\photon-3B}
& 0.247
& 0.360
& 0.455\\

\rowcolor{lightgray} \textbf{\photon-1B}
& 0.243
& 0.215
& 0.349\\
\end{tabular}
}
\label{tab:eval_gauntlet_1}
\end{table}

\begin{table}[t]
\caption{In-context learning comparison between \photon models. Our biggest model wins $3$ out of $3$ comparisons in this group.}
\resizebox{\columnwidth}{!}{
\begin{tabular}{@{}lccc@{}}
\toprule

\textbf{Name}
& \begin{tabular}[c]{@{}c@{}}\textbf{PIQA} \\ \citep{piqa}\end{tabular}
& \begin{tabular}[c]{@{}c@{}}\textbf{Winogrande} \\ \citep{winogrande}\end{tabular}
& \begin{tabular}[c]{@{}c@{}}\textbf{ARC-Easy}\\ \citep{arc_challenge}\end{tabular}\\
\midrule
\rowcolor{lightgray} \textbf{\photon-7B}
& $\mathbf{0.729}$
& $\mathbf{0.522}$
& $\mathbf{0.508}$\\

\textbf{\photon-3B}
& 0.705
& 0.512
& 0.461\\

\rowcolor{lightgray} \textbf{\photon-1B}
& 0.676
& 0.516
& 0.390\\
\end{tabular}
}
\label{tab:eval_gauntlet_2}
\end{table}

The takeaways of this evaluation are: (1) federated training converges for all dropout ratios $r < 0.875$, making it suitable for highly unreliable hardware configurations, (2) since nodes train in isolation, a node failure does not require interrupting the entire federated round, rather it only reduces the number of pseudo-gradients used for an update, (3) to compensate for such failures, it is sufficient to extend training until the target number of tokens is reached, and (4) configurations with higher dropout ratios correspond to a reduction in the effective batch size of the training, which may improve the final performance at the cost of longer training times.

\subsection{Downstream evaluation of \photon's models}\label{app_sec:downstrea_eval}

To evaluate the downstream task performance of our models, we test across a series of in-context learning benchmarks. Our results, shown in \cref{tab:eval_gauntlet_1,tab:eval_gauntlet_2,tab:eval_gauntlet_3,tab:eval_gauntlet_4}, demonstrate that the downstream performance of models trained with \photon scales as expected with model size, with our largest model winning $10$ out of $14$ comparisons. This proves the downstream utility of \photon models even when using a pre-training dataset not optimized for downstream performance. We expect that as we increase the model size and incorporate a broader and more qualitative data mixture, the downstream performance of \photon models will keep improving.
\begin{table}[H]
\caption{In-context learning comparison between \photon models. Our biggest model wins $2$ out of $3$ comparisons in this group.}
\resizebox{\columnwidth}{!}{
\begin{tabular}{@{}lccc@{}}
\toprule

\textbf{Name}
& \begin{tabular}[c]{@{}c@{}}\textbf{BoolQ} \\ \citep{boolq}\end{tabular}
& \textbf{\begin{tabular}[c]{@{}c@{}}\textbf{Openbook QA} \\ \citep{openbookqa}\end{tabular}}
& \textbf{\begin{tabular}[c]{@{}c@{}}\textbf{Winograd} \\ \citep{winograd}\end{tabular}} \\
\midrule
\rowcolor{lightgray} \textbf{\photon-7B}
& 0.530
& $\mathbf{0.358}$
& $\mathbf{0.681}$\\

\textbf{\photon-3B}
& 0.591
& 0.316
& 0.656\\

\rowcolor{lightgray} \textbf{\photon-1B}
& $\mathbf{0.612}$
& 0.274
& 0.604\\
\end{tabular}
}
\label{tab:eval_gauntlet_3}
\end{table}

\begin{table}[H]
\caption{In-context learning comparison between \photon models. Our biggest model wins $3$ out of $4$ comparisons in this group.}
\resizebox{\columnwidth}{!}{
\begin{tabular}{@{}lcccccc@{}}
\toprule
\textbf{Name}
& \textbf{\begin{tabular}[c]{@{}c@{}}LAMBADA (OpenAI) \\ \citep{lambada}\end{tabular}}
& \textbf{\begin{tabular}[c]{@{}c@{}}Bigbench Strategy QA\\ \citep{bigbench}\end{tabular}}
& \textbf{\begin{tabular}[c]{@{}c@{}}\textbf{COPA} \\ \citep{copa}\end{tabular}}
& \textbf{\begin{tabular}[c]{@{}c@{}}\textbf{MMLU} \\ \citep{mmlu}\end{tabular}} \\ \midrule
\rowcolor{lightgray}  \textbf{\photon-7B}
& $\mathbf{0.457}$
& 0.466
& $\mathbf{0.710}$
& $\mathbf{0.263}$ \\
\textbf{\photon-3B}
& 0.381
& 0.464
& 0.620
& 0.252 \\
\rowcolor{lightgray}  \textbf{\photon-1B}
& 0.298
& $\mathbf{0.470}$
& 0.630
& 0.248 \\
\end{tabular}
}
\label{tab:eval_gauntlet_4}
\end{table}
\FloatBarrier
\clearpage
{\onecolumn
\section{Full Algorithms} \label{app_subsec:full_algo}

\begin{algorithm}[ht]
\caption{Distributed Data Parallel (DDP) Training Algorithm}
\label{alg:ddp}
\scalebox{0.95}{
\begin{minipage}{0.95\linewidth}
\begin{algorithmic}[1]
\Require{$N$: Number of devices (workers), $f_{\theta}$: Model with parameters $\theta$,$T$: Number of epochs}
\Require{$\mathcal{D}$: Dataset partitioned across devices $\mathcal{D}_i$ where $i \in \{1, 2, \dots, N\}$}
\Require{$\texttt{RingAllReduce}$: All-reduce operation to aggregate across devices on a ring}
\Require{$\texttt{Opt}$: Optimizer for updating $\theta$ with gradients}

\State{\textbf{Initialize:} Randomly initialize model parameters $\theta_0$ on each device}
\For{$t = 1$ to $T$}
    \State \textbf{Step 1: Parallel Local Training}
    \For{each device $i \in \{1, 2, \dots, N\}$ \textbf{in parallel}}
        \State Compute local mini-batch loss $\mathcal{L}_i(\theta_{t-1}, \mathcal{D}_i)$
        \State Compute local gradients $\nabla_{\theta_{t-1}} \mathcal{L}_i(\theta_{t-1})$
    \EndFor
    \State \textbf{Step 2: Distributed \texttt{RingAllReduce} Gradient Aggregation}
    \State $\nabla_{\theta_{t-1}} \mathcal{L} = \frac{1}{N} \sum_{i=1}^N \nabla_{\theta_{t-1}} \mathcal{L}_i$
    \State \textit{Each device} now possesses the global gradient $\nabla_{\theta_{t-1}} \mathcal{L}$
    \State \textbf{Step 3: Parallel Model Update}
    \For{each device $i \in \{1, 2, \dots, N\}$ \textbf{in parallel}}
        \State $\theta_t = \texttt{Opt}(\theta_{t-1}, \nabla_{\theta_{t-1}} \mathcal{L})$
    \EndFor
\EndFor
\State{ \textbf{Output:} Trained model parameters $\theta_T$}
\end{algorithmic}
\end{minipage}
}
\end{algorithm}

\begin{algorithm}[hb]
\caption{Cross-silo Federated Learning (FL) Algorithm}
\label{alg:cross_silo_fl}
\scalebox{0.95}{
\begin{minipage}{0.95\linewidth}
\begin{algorithmic}[1]
\Require{$N$: Number of clients, $f_{\theta}$: Model with parameters $\theta$}
\Require{$T$: Number of federated rounds, $K$: Number of local steps}
\Require{$\{\mathcal{D}_i\}$: Federated dataset, i.e., a set of private $\mathcal{D}_i$, $i \in \{1, \dots, N\}$}
\Require{$\texttt{ClientOpt}$: local client optimizer, $\texttt{ServerOpt}$: server optimizer}

\State{ \textbf{Initialize:} Randomly initialize global model parameters $\theta_0$ on the server}

\For{$t = 1$ to $T$}
    \State \textbf{Step 1: Broadcast model parameters}
    \State Server sends $\theta_t$ to all $N$ clients
    \State \textbf{Step 2: Parallel Local Training}
    \For{each client $i \in \{1, 2, \dots, N\}$ \textbf{in parallel}}
        \State $\omega_{i,0} \gets \theta_{t-1}$
        \For{each local iteration $k \in \{1, 2, \dots, K\}$}
            \State Compute local mini-batch loss $\mathcal{L}_i(\omega_{i,k-1}, \mathcal{D}_i)$
            \State Compute local gradients $\nabla_{\omega_{i, k-1}} \mathcal{L}_i(\omega_{i,k-1})$
            \State $\omega_{i, k} \gets \texttt{ClientOpt}(\omega_{i, k-1}, \nabla_{\omega_{i, k-1}} \mathcal{L}_i(\omega_{i, k-1}))$
        \EndFor
        \State $\Delta\theta_{t-1,i} \gets \omega_{i, K} - \theta_{t-1}$
    \EndFor
    \State \textbf{Step 3: Global Model Update (on the server)}
    \State $\theta_t = \texttt{ServerOpt}(\theta_{t-1}, \{\Delta\theta_{t-1,i}\})$
\EndFor
\State \textbf{Output:} Trained model parameters $\theta_T$
\end{algorithmic}
\end{minipage}
}
\end{algorithm}

}
\twocolumn
\FloatBarrier
\clearpage
\section{Errata Corrige: Corrected MFU Reporting}
\label{app_sec:errata_corrige}

After publication, we identified an error in the Model FLOPs Utilization (MFU)
values reported for the billion-scale system metrics in
\cref{tab:federation_scalability}. The affected runs were executed on NVIDIA
H100 80GB HBM3 GPUs, but the historical logging configuration normalized
\texttt{throughput/\allowbreak device/\allowbreak flops\_per\_sec} by an
A100-style dense FP16/BF16 peak of
$312 \times 10^{12}$ FLOPs/s. For H100 HBM3 under the same dense FP16/BF16
Composer convention, the correct per-device peak is
$989.5 \times 10^{12}$ FLOPs/s.

We corrected the MFU values by preserving the original device FLOPs/s and
renormalizing by the H100 dense FP16/BF16 peak:
\[
    \mathrm{MFU}_{\mathrm{corrected}} =
    \frac{\mathrm{device\ FLOPs/s}}{989.5 \times 10^{12}}.
\]
Historical W\&B run records retain their original raw
\texttt{throughput/\allowbreak device/\allowbreak mfu} metric for traceability.
Corrected values are recorded separately under
\texttt{throughput/\allowbreak device/\allowbreak mfu\_corrected} by the audit
workflow.

\begin{table}[h]
\caption{\textbf{MFU corrections for \cref{tab:federation_scalability}.}
Centralized rows were recomputed from W\&B device FLOPs/s. Federated rows use
the deterministic denominator correction from the originally reported table
value when the exact source run could not be uniquely identified.}
\label{tab:mfu_errata_corrige}
\vspace{0.1cm}
\centering
\small
\begin{tabular}{@{}lcc@{}}
\toprule
\textbf{Model} & \textbf{Original} & \textbf{Corrected} \\
\midrule
Cen-$1.3$B & $0.8027$ & $0.2531$ \\
Fed-$1.3$B & $1.1245$ & $0.3546$ \\
Cen-$3$B & $0.165$ & $0.051$ \\
Fed-$3$B & $0.240$ & $0.076$ \\
Cen-$7$B & $0.335$ & $0.105$ \\
Fed-$7$B & $0.224$ & $0.071$ \\
\bottomrule
\end{tabular}
\vspace{-0.2cm}
\end{table}

This correction affects only the reported MFU column. Perplexity, wall time,
compute time, communication time, GPU utilization, and the relative conclusions
from the system comparison are unchanged.

\section{Artifact Appendix}

\subsection{Abstract}
This Artifact Appendix provides the instructions, scripts, and configurations necessary to run the experiments of our paper on federated large language model (LLM) pre-training using the \textit{Photon} system.
We focus on the script, \texttt{scripts/fed\_125m\_example.sh}, that orchestrates the entire process: downloading dependencies, launching the federated server, spinning up clients, and training a 125M-parameter model end to end.
However, we recommend following carefully the \texttt{README.md} file and the provided example scripts for a more detailed understanding of the setup and execution.
By running the \texttt{scripts/fed\_125m\_example.sh} script, users can witness how Photon handles Hydra-based configuration resolution, aggregator (server) bootstrapping, and client participation.

\subsection{Artifact check-list (meta-information)}
{\small
\begin{itemize}
  \item \textbf{Algorithm:} LocalSGD-based federated optimization with integrated distributed data-parallel (DDP) or fully sharded data parallel (FSDP) when applicable.
  \item \textbf{Program:} Python scripts employing PyTorch, integrated with Flower (for federated coordination) and Ray for model updates communication.
  \item \textbf{Compilation:} No explicit compilation. A Python-based environment setup is mandatory.
  \item \textbf{Transformations:} Data tokenization, normalization, optional data pre-processing (compression), and partitioning in client shards.
  \item \textbf{Binary:} No direct binaries; entire artifact is Python-based.
  \item \textbf{Data set:} A small subset of C4 is included for demonstration. For larger training, full C4 or The Pile can be substituted (scripts not included here).
  \item \textbf{Run-time environment:} Linux system (Ubuntu 22.04 recommended), Python 3.11, CUDA(12.4)-enabled PyTorch 2.1.5, plus Hydra for configuration resolution.
  \item \textbf{Hardware:} At least one NVIDIA GPU (NVIDIA A40, RTX2080Ti, V100, A100, H100, etc.), stable network links (1--10Gbps) if multiple machines are used.
  \item \textbf{Run-time state:} Users can run everything on a single machine with multiple GPUs, or distribute across multiple nodes.
  \item \textbf{Execution:} A single script \texttt{scripts/fed\_125m\_example.sh} that performs the entire flow (setup, server launch, client launches, local training).
  \item \textbf{Metrics:} Primary metric is validation perplexity, with secondary metrics including GPU utilization, throughput, and communication overhead. Wandb logging is supported but requires custom configuration for which guidelines are provided in the code docstrings.
  \item \textbf{Output:} Model checkpoints, logs of training progress, final perplexity.
  \item \textbf{Experiments:} Demonstration of the federated pre-training and centralized training of a $125$M-parameter decoder-only LLM, which can be scaled up if desired.
  \item \textbf{Disk space required:} Approximately $5/15$GB for the small subset of C4 plus checkpoints. (Larger experiments may require $300/1000$GB).
  \item \textbf{Time needed to prepare workflow:} Approximately 1 hour for environment setup, $30/60$ minutes to download and preprocess the small dataset.
  \item \textbf{Time needed to complete experiments:} A few hours for the $125$M demonstration. Larger-scale runs can take days.
  \item \textbf{Publicly available:} Yes, code repository is licensed (Apache-2.0 license) and will be made public.
  \item \textbf{Code licenses (if publicly available):} Apache License 2.0.
  \item \textbf{Data licenses (if publicly available):} C4 is under the ODC-BY license.
  \item \textbf{Workflow framework used:} Flower + Ray + PyTorch + Hydra, plus a single orchestrating shell script.
  \item \textbf{Archived:} \href{https://doi.org/10.5281/zenodo.15187915}{DOI on Zenodo}.
  \item \textbf{Public permalink:} \href{http://flower.ai/research}{Flower Labs Research}.
\end{itemize}
}

\subsection{Description}

\subsubsection{How delivered}
The artifact is provided in a zipped repository containing:
\begin{itemize}
  \item \texttt{README.md}: A quick overview and key instructions.
  \item \texttt{scripts/system\_setup.sh}: Installs base dependencies, sets up the environment.
  \item \texttt{scripts/convert\_c4\_dataset.sh}:  Acquires a small version of C4 for demonstration. Prepare the dataset for training.
  \item \texttt{scripts/fed\_125m\_example.sh}:\\The single script that launches everything for a 125M-parameter model. It internally invokes Hydra-based configs for server and clients, then orchestrates the run.
  \item \texttt{scripts/cen\_125m\_example.sh}:\\The single script that launches centralized training of a 125M-parameter model. It internally invokes Hydra-based configs. It is prepared to operate on a single machine setup launching a parallelized training on the available GPUs.
  \item \texttt{configs/}: YAML files specifying hyperparameters (learning rate, batch size, etc.), aggregator properties, and Hydra overrides.
\end{itemize}

\subsubsection{Hardware dependencies}
\begin{itemize}
  \item \textbf{GPU:}
    \begin{itemize}
      \item For the 125M example, a single GPU with \(\geq\)12GB memory is sufficient, even though a larger memory (\(\geq\)40GB) is recommended.
      \item For multi-node, each node should have a CUDA-capable GPU and at least 1--10Gbps network connectivity.
    \end{itemize}
\end{itemize}

\subsubsection{Software dependencies}
\begin{itemize}
  \item \textbf{OS:} Linux (Ubuntu 22.04+).
  \item \textbf{Python:} 3.11 or higher.
  \item \textbf{CUDA/CuDNN:} Version 12.4 is recommended, being compatible with PyTorch 2.1.5 and your specific GPU driver. These can be installed automatically via \texttt{scripts/system\_setup.sh}
  \item \textbf{Package managers:} Poetry is supported for dependency management.
  \item \textbf{Libraries:} PyTorch 2.1.5, Flower (custom version), Ray, Hydra, and standard Python utilities (NumPy, Pandas, etc.). Installed automatically via the \texttt{scripts/system\_setup.sh} and \texttt{scripts/install\_env.sh} scripts.
\end{itemize}

\subsubsection{Data sets}
\begin{itemize}
  \item A small subset of C4 is included for demonstration.
  \item It is fetched, unpacked locally, and tokenized by\\ \texttt{scripts/convert\_c4\_dataset.sh}.
\end{itemize}
Users can later replace this with the full C4 or other corpora by adjusting parts of the code and configuration files.

\subsection{Installation and Usage}

Refer to the \texttt{README.md} file for a more detailed guide. Below is a quick start guide to run the federated pre-training of a 125M-parameter model.

\noindent
\textbf{System prep and environment:}
\begin{enumerate}
  
  \item \textbf{Download the code:} The code is maintained and made available through the \href{https://flower.ai/research}{Flower Labs Research} webpage.
  \item \textbf{Run the setup script:} Once the repository has been obtained, run the setup script to install the necessary dependencies and prepare the environment.
  \begin{verbatim}
cd <path>/<to>/<photon>
cd scripts
. system_setup.sh
  \end{verbatim}
  This can install build tools, CUDA drivers (Ubuntu-based).
  \item \textbf{Install dependencies:}
  \begin{verbatim}
cd scripts
. install_env.sh
  \end{verbatim}
\end{enumerate}
\textbf{Download, prepare/convert dataset with the provided script.}
\begin{verbatim}
bash scripts/convert_c4_dataset.sh
\end{verbatim}
\noindent
\textbf{Run the single script for federate pre-training of the 125M model:}
\begin{verbatim}
bash scripts/fed_125m_example.sh
\end{verbatim}
This command executes the following steps internally:
\begin{itemize}
  \item \textbf{Hydra configs interpretation:} Hydra interprets the configs and dumps them to a file that is read by the other processes. The file \texttt{photon/hydra\_resolver.py} is used.
  \item \textbf{Launch Flower Superlink:} The command used is \texttt{poetry run flower-superlink}.
  \item \textbf{Launch Flower ServerApp:} The command used is \texttt{poetry run flower-server-app photon.server\_app:app}.
  \item \textbf{Launch Flower ClientApps:} The command used is \texttt{poetry run flower-client-app photon.client\_app:app}
  \item \textbf{Federated rounds:} The aggregator orchestrates local training (LocalSGD) across clients, synchronizes updates after each round.
  \item \textbf{Checkpoints and logs:} Intermediate global checkpoints and logs are saved in \texttt{checkpoints/} and \texttt{runs/} respectively.
  \item \textbf{Completion:} The script logs periodically several metrics, e.g., perplexity and throughput.
\end{itemize}

\subsection{Evaluation and expected result}
\textbf{Targets of interest:}
\begin{itemize}
  \item \textbf{Validation perplexity:} For the 125M demo, you should observe perplexity dropping towards the low 40s or upper 30s after sufficient rounds, depending on configuration.
  \item \textbf{Runtime logs:} Both aggregator and client logs are found under \texttt{runs/}, indicating the number of tokens processed, average GPU utilization, and steps per round.
  \item \textbf{Checkpoints:} Partial and final checkpoints are saved in the\\ \texttt{checkpoints/} folder.
\end{itemize}

\subsection{Experiment customization}
\begin{itemize}
  \item \textbf{Config override:} Edit \texttt{scripts/fed\_125m\_example.sh} or pass Hydra overrides to change client count or hyperparameters.
  \item \textbf{Hardware scaling:} By default, the script spawns multiple clients on a single node. For multi-node, adapt the aggregator IP and client addresses in \texttt{scripts/fed\_125m\_example.sh}.
  \item \textbf{Batch sizes / epochs:} Controlled by Hydra configs in the\\ \texttt{configs/} folder.
  \item \textbf{Dataset:} Replace the small C4 path with your own local data for more extended training.
\end{itemize}

\subsection{Notes}
\begin{itemize}
  \item \textbf{Partial or intermittent clients:} If a client crashes or is not reachable, the aggregator  continues with remaining clients in subsequent rounds.
  \item \textbf{Performance considerations:} For minimal overhead, ensure a stable GPU environment. Larger-scale runs (1.3B+) require more disk space, memory, and  multi-GPU setups.
\end{itemize}

\subsection{Methodology}
\noindent
We adhere to artifact evaluation guidelines:
\begin{itemize}
  \item Single-blind AE with emphasis on reproducibility and clarity.
  \item Clear \textit{build} (ffrom the scripts \texttt{scripts/system\_setup.sh} and\\ \texttt{scripts/install\_env.sh}), \textit{run} (using the script\\ \texttt{scripts/fed\_125m\_example.sh}), and \textit{analysis} (logs, final checkpoint) phases.
  \item \textbf{ACM Artifact Badging} \href{https://www.acm.org/publications/policies/artifact-review-badging}{best practices}: code will be made public, well-documented, and tested on a standard environment.
\end{itemize}

\end{document}